\pdfoutput=1

\documentclass[11pt]{article}

\usepackage[final]{acl}

\usepackage{times}
\usepackage{latexsym}
\usepackage{subcaption}

\usepackage{booktabs}

\usepackage[T1]{fontenc}

\usepackage[utf8]{inputenc}

\usepackage{microtype}

\usepackage{inconsolata}

\usepackage{graphicx}
\usepackage{enumitem}

\title{A Rigorous Evaluation of LLM Data Generation Strategies\\for Low-Resource Languages}

\author{Tatiana Anikina$^{\spadesuit}$\thanks{Equal contribution.}, Jan Cegin$^\dagger$$^\ddagger$\footnotemark[1], Jakub Simko$^\ddagger$, Simon Ostermann$^{\spadesuit}$ \\
  $^{\spadesuit}$ German Research Institute for Artificial Intelligence (DFKI), Saarbr\"{u}cken, Germany \\
    \texttt{\{tatiana.anikina, simon.ostermann\}}@dfki.de \\
  $^\dagger$ Faculty of Information Technology, Brno University of Technology, Brno, Czechia \\
  $^\ddagger$ Kempelen Institute of Intelligent Technologies, Bratislava, Slovakia\\
  \texttt{\{jan.cegin, jakub.simko\}}@kinit.sk \\}

\begin{document}
\maketitle
\begin{abstract}
Large Language Models (LLMs) are increasingly used to generate synthetic textual data for training smaller specialized models. However, a comparison of various generation strategies for low-resource language settings is lacking.  While various prompting strategies have been proposed—such as demonstrations, label-based summaries, and self-revision—their comparative effectiveness remains unclear, especially for low-resource languages. In this paper, we systematically evaluate the performance of these generation strategies and their combinations across 11 typologically diverse languages, including several extremely low-resource ones. Using three NLP tasks and four open-source LLMs, we assess downstream model performance on generated versus gold-standard data. Our results show that strategic combinations of generation methods—particularly target-language demonstrations with LLM-based revisions—yield strong performance, \textbf{narrowing the gap with real data to as little as 5\% in some settings}. We also find that smart prompting techniques can \textbf{reduce the advantage of larger LLMs}, highlighting efficient generation strategies for synthetic data generation in low-resource scenarios with smaller models.
\end{abstract}

\section{Introduction}

\begin{figure}[t!]
    \centering
    \includegraphics[width=5cm]{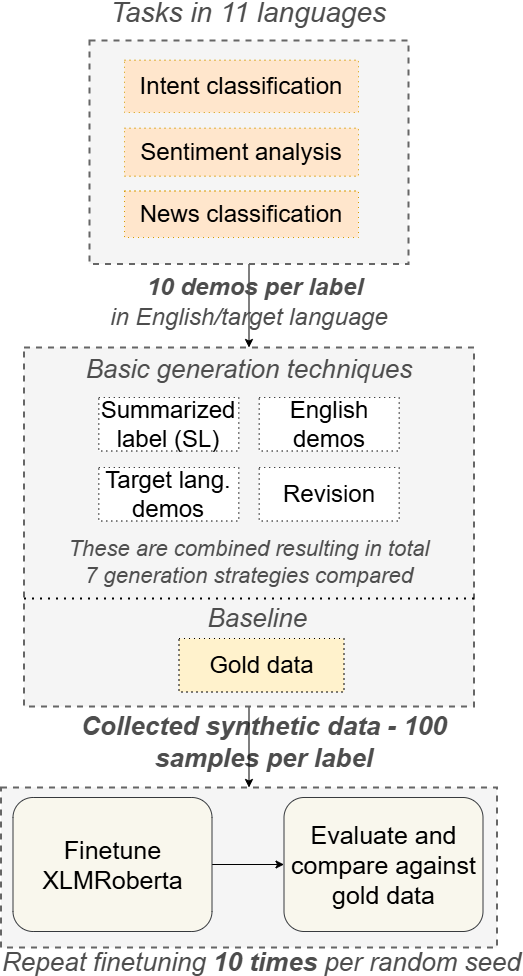}
    \caption{Overview of our methodology. We consider 11 different languages (including English) for 3 different tasks. We use various generation strategies to create synthetic data for each strategy, language and LLM combination. XLM-R is then finetuned and evaluated on the test set of the original dataset for the given task and language.}
    \label{fig:methodology}
\end{figure}

Large language models (LLMs) such as GPT-4, Gemini, and Llama show impressive performance in producing well-formed text - trivially. This essential capability makes them ideal for both the \emph{augmentation} and \textit{generation} of text datasets~\cite{ubani2023zeroshotdataaug, dai2023auggpt, piedboeuf-langlais-2023-chatgpt, cegin-etal-2023-chatgpt, cegin-etal-2025-llms}. Such LLM-based text generation is often leveraged for creating representative data that are used for training smaller, more efficient downstream models, a technique that is sometimes referred to as \textit{LLM distillation} \cite{xu2024survey}. LLM-based text generation has been used for various downstream tasks such as sentiment analysis~\cite{ONAN2023101611, piedboeuf-langlais-2023-chatgpt}, intent recognition~\cite{cegin2024userandomselectionnow} and news topic classification~\cite{piedboeuf-langlais-2023-chatgpt, cegin2024effectsdiversityincentivessample}. While most studies focus only on English, LLMs have also been leveraged for textual generation in low-resource languages in some cases~\cite{namboori2023gemquad, glenn-etal-2023-jetsons, ZOTOVA2021114547, chung2025beyond}, showing impressive performance in generating textual data for a wide variety of languages 
and often outperforming neural machine translation as augmentation technique~\cite{pranida2025syntheticdatagenerationculturally}.

Various generation strategies exist for generating high-quality textual data, such as including textual examples of the desired data in the generation prompt itself~\cite{cegin2024effectsdiversityincentivessample}, using descriptions of the labels of data to be generated in the generation prompt~\cite{xu2024-etal, abraham2025promptselectionmattersenhancing} and using self-revision of the LLMs after generation to filter generated data~\cite{li-etal-2024-llatrieval}. \textbf{However, two essential gaps remain}. First, the performance of such generation strategies is often evaluated in isolation, and their combinations have not been compared, leaving the whole potential of LLM-based data augmentation obscure. Second, they have been used for generating mostly English texts, which leaves most languages underexplored - especially low-resource ones. To the best of our knowledge, an overview of which generation strategies (or their combination) are best for textual generation in low-resource languages is completely {lacking}.

The goal of this paper is to close this gap and compare existing generation strategies for low-resource text generation in various languages. Our methodology is shown in Figure \ref{fig:methodology}. We evaluate the performance of strategies based on the downstream task performance, using smaller models finetuned on the generated data. We perform a comparative analysis of different generation strategies using 11 typologically diverse languages (including English) with different scripts, including several very low-resource languages such as Welsh, Romanian, Azerbaijani, Slovenian, and Telugu. As generation strategies, we consider 3 prominent approaches:
\textit{Summarized label} - using explanations of the data to be generated~\cite{xu2024-etal, abraham2025promptselectionmattersenhancing}; \textit{Demonstrations} - the inclusion of examples in the target language or English in the prompts~\cite{piedboeuf-langlais-2023-chatgpt, cegin2024effectsdiversityincentivessample}; And \textit{Revision} - additionally filtering out samples that are generated using either of the previous two strategies, using the same LLM~\cite{li-etal-2024-llatrieval}. 
We benchmark four open-source LLMs of different sizes on three NLP tasks and perform an ablation study with respect to the language of demonstrations, intent description, and LLM-based revision. We measure the performance of downstream models for various generation strategies and their combinations. For each task, we compare the configurations to an upper bound that is set by the same model being trained on the same amount of real data. All generated data will be released upon acceptance and our code is available at \url{https://github.com/tanikina/multilingual-generation}.


Our main contributions are:
\begin{itemize}
    \item We provide an exhaustive evaluation of different common strategies for \textbf{generating synthetic data for low-resource languages} and formulate suggestions for the most effective combination of strategies for extreme low-resource settings.
    \item We confirm that while models with a higher number of parameters outperform their smaller counterparts on the generation task in the low-resource setting, \textbf{the gap in performance is small} when a right generation technique is used.
    \item We show that - using the right combination of techniques, and for some configurations - \textbf{the drop of performance for a model trained on LLM-generated data is as small as up to only 5\%} absolute when compared to fine-tuning on the same amount of ``real'' data. 
    \item We show that a \textbf{combination of demonstrations in the target language with LLM-based revisions} generally leads to the best performance across most languages, especially in extremely low-resource settings.
\end{itemize}

\section{Related Work}
Soon after their advent, new LLMs, such as GPT-4 or Llama started to be used as data augmentation or data generation tools. In general, this newly generated data is used to train smaller downstream models, for better efficiency. LLM-based augmentation is typically done through paraphrasing~\cite{cegin2024effectsdiversityincentivessample, dai2023auggpt, sen-etal-2023-people, evuru2024coda}. Less often, LLMs are used to create semantically new samples adhering to a given label~\cite{ubani2023zeroshotdataaug, cegin2024userandomselectionnow}. LLM-based augmentation and data generation has been used for a variety of tasks such as automated scoring~\cite{fang2023using}, low-resource language generation~\cite{ghosh-etal-2023-dale}, intent classification~\cite{sahu-etal-2022-data}, sentiment analysis ~\cite{piedboeuf-langlais-2023-chatgpt, ONAN2023101611, yoo-etal-2021-gpt3mix-leveraging}, hate speech detection~\cite{sen-etal-2023-people}, news classification~\cite{piedboeuf-langlais-2023-chatgpt}, content recommendation~\cite{contect-based-recom}, and health symptoms classifications~\cite{dai2023auggpt}.

While the main focus in previous research is on creating textual data for English, some works have also explored synthetic data generation for other languages. Machine Translation-based techniques leverage multilingual models for enhancing textual data in languages such as Vietnamese~\cite{feng-etal-2021-survey}. Variants of this approach use chain-of-thought prompting with LLMs~\cite{son2025efficient}. Multilingual synthetic generation using LLMs has been leveraged for various tasks like QA~\cite{kramchaninova-defauw-2022-synthetic, namboori2023gemquad}, fact-checking~\cite{chung2025beyond}, NER~\cite{liu-etal-2021-mulda}, sentiment stance detection~\cite{ZOTOVA2021114547} and classification~\cite{glenn-etal-2023-jetsons} - while not focusing on low-resource languages particularly. A study comparing human-created data, machine translation-created data, and LLM-generated synthetic data for culturally nuanced commonsense reasoning in low-resource languages has found that using LLM-generated data can be better than machine translation for downstream classification~\cite{pranida2025syntheticdatagenerationculturally}. 

While the usage of LLM-based data generation is growing for non-English languages, a comparison of how such synthetic data should be generated, especially for low-resource languages, is lacking.


\section{Methodology and Experiments}

Our methodology is visualized in Figure~\ref{fig:methodology} and follows a simple procedure: For a given task, we extract 10 demos per label in a target language. We then use a range of generation strategies to collect 100 synthetic data points per label. Finally, we train an XLM-R model \cite{DBLP:journals/corr/abs-1911-02116} on the data. Our chose XLM-R because it represents an encoder-only, efficient and small model (279M parameters) that supports 94 languages. We also performed pilot experiments with mBERT \cite{devlin-etal-2019-bert} but found that XLM-R achieves better scores on average, therefore we focused on the XLM-R downstream model for the final evaluation.

In this project we follow a controlled setting with the same amount of samples per label, i.e. 10 labels for the intent recognition task results in 1,000 synthetic samples, 7 labels for topic classification in 700 samples, and 2 labels for sentiment analysis in 200 samples. In order to address potential overfitting, we use early stopping with a patience of 5 epochs and although our training and development data are from the generated pool, all test sets are always from the original gold dataset, which prevents the model from performing too well by memorizing training samples. To ensure a robust comparison of various generation strategies, we use 11 different languages (including English), 3 tasks and 4 different LLMs of varying sizes. 



\subsection{Generation Strategies}\label{sec:gen-strategies}

We evaluate three strategies: \textit{Summarized Label}, \textit{Demonstrations}, and \textit{Revision}, which are outlined below.


\paragraph{Summarized Label.} For this strategy, we generate label descriptions by prompting Llama3-70b with 10 English samples per label (see Figure \ref{fig:summarized_label_template} for the prompt and Section \ref{sec:summarized_intent_examples} for the summarized label examples). The generated intent summary is then added to the final prompt. Although many labels in the datasets have self-descriptive names (e.g. \textit{geography}, \textit{cooking\_recipe}), adding a summarized description may provide additional information that is useful for text generation.

\paragraph{Demonstrations.} We compare two settings: Using  English examples of the target label (from which we want to generate samples) or using target language examples of the target label. This way we can assess the impact of the target language demonstrations and simulate a scenario when resources for non-English languages are unavailable. 
To simulate a low-resource scenario for all cases, we limit the number of demonstrations to 10. We provide only the demonstrations from the class we want to generate and select them randomly, following~\cite{cegin2024userandomselectionnow}. 

\begin{figure*}[h!]
    \centering
    \includegraphics[width=0.7\textwidth]{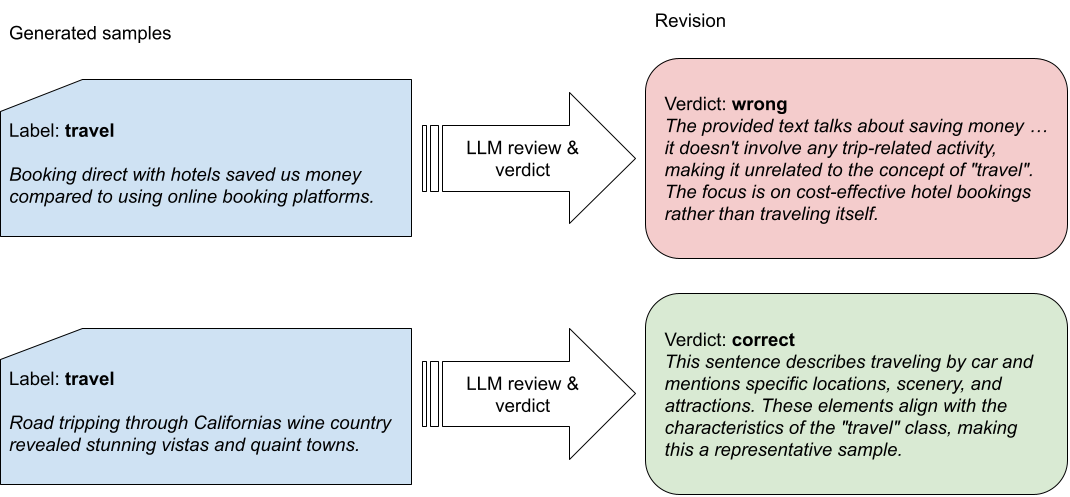}
    \caption{LLM-based revision examples.}
    \label{fig:revision-example}
\end{figure*}

\paragraph{Revision.} We use an LLM to revise and filter out ``bad examples'' by prompting with the intent description and previously generated samples. This strategy is used in combination with the previous two, as it requires generated data to be used. Figure \ref{fig:revision-example} provides an example of samples that were accepted or rejected by the judge LLM. Figure \ref{fig:revision-prompt} shows the prompt for LLM-based assessment of the generated samples.

\paragraph{Additional Combinations.}
We also consider two additional combinations of the generation strategies: We combine the summarized label descriptions with demonstrations in English and the target language without revision in both cases. As \textit{revision} has the highest cost of all strategies (additional inference of the LLM), its exclusion in these combinations enables us to measure what impact on downstream model performance revision has.


\subsection{Data and Tasks}\label{sec:data-tasks}
We consider three different tasks: intent recognition, topic classification, and sentiment analysis. We generate artificial data for each of these tasks in 11 typologically diverse languages using the generation strategies outlined above. We use 2 high-resource languages, German and English, as a comparison. Our experiments include mid-resourced languages: Thai, Hebrew, Indonesian, and Swahili; as well as low-resourced ones: Romanian, Azerbaijani, Slovenian, Telugu, and Welsh. This selection of languages is based on their diversity and the availability of data for each of the tasks.

For intent recognition, we use MASSIVE \cite{fitzgerald-etal-2023-massive}, which is a multilingual dataset for virtual assistant evaluation with intent and slot annotations that supports 51 languages and 60 intents. Given that our focus on data generation in low-resource settings introduces additional complexities, we simplify the intent recognition task by considering only ten distinct labels from the original dataset (see \ref{app:massive-examples} for the examples). 


For topic classification, we chose SIB-200 \cite{adelani-etal-2024-sib} that has 7 labels (see \ref{app:sib200-examples} for the examples) and also supports a variety of high and low-resource languages. SIB-200 is based on the Flores-200 machine translation corpus, and it uses sentence-level annotations.

As the third task, we selected sentiment classification. To the best of our knowledge, there is currently no unified dataset available for this task that contains a wide range of high and low-resource languages. Thus, in our experiments, we consider 10 different datasets for low-resource languages from \cite{gurgurov-etal-2025-gremlin} and \cite{gurgurov-etal-2024-adapting} and two additional datasets for English and German from \cite{mollanorozy-etal-2023-cross}. Note that these datasets substantially vary in terms of topic coverage and text format. For example, the German dataset contains mostly sentiment with respect to transportation and infrastructure, while the Romanian dataset is about product reviews. We acknowledge that although all samples in these datasets express positive or negative sentiment, some topics might be an easier target for text generation.

\subsection{Models}\label{sec:models}

We generate artificial samples with 4 models of different sizes: Gemma-3 \cite{gemma_2025} with 4 and 27 billion parameters, and Llama-3 \cite{llama3modelcard} with 8 and 70 billion parameters. These models were chosen for their open-source nature and support for multiple languages. Although the extent to which each language was represented in the training data is unclear, we were able to generate samples for all 11 target languages.

\subsection{Experiments}\label{sec:experiments}

We generate 100 samples per label for each dataset and language using the generation strategies described in \ref{sec:gen-strategies} (see \ref{sec:computational_resources} for the details on computational resources). After collecting the data, we fine-tuned XLM-R~\cite{DBLP:journals/corr/abs-1911-02116} on generated data for all three tasks and eleven languages. As an upper bound for comparison, we also train the model on the same amount of balanced gold data, i.e.~100 examples per class. We fine-tune and evaluate 10 models for each configuration and report macro F1 scores averaged across 10 runs. Details regarding the downstream model fine-tuning can be found in \ref{sec:downstream_finetuning}.

\begin{table*}[h!]
\resizebox{\textwidth}{!}{
\begin{tabular}{@{}llllllllllll|l@{}}
\toprule
 \textbf{Intent Recognition} & az & cy & de & en & he & id & ro & sl & sw & te & th & avg \\ \midrule
Gold Data & \textit{91.62} & \textit{91.46} & \textit{94.38} & \textit{95.32} & \textit{92.17} & \textit{92.78} & \textit{94.27} & \textit{93.66} & \textit{89.75} & \textit{90.62} & \textit{94.27} & \textit{92.76} \\\hline
Summarized Label (SL) & 78.13 & 63.22 & 88.06 & 92.42 & 78.54 & 90.00 & 89.01 & 88.47 & 71.85 & 81.85 & 86.35 & 82.53 \\
EnglishDemos + SL & 82.25 & 68.96 & 92.08 & \underline{92.79} & 83.53 & 91.30 & 89.22 & 85.99 & 71.06 & 79.66 & 86.26 & 83.92 \\
EnglishDemos + Rev. & 83.72 & 67.93 & \underline{92.40} & \textbf{93.03} & 84.10 & \underline{91.73} & 90.50 & 87.18 & 71.87 & 83.12 & 86.83 & 84.77 \\
TargetDemos & 79.91 & 68.45 & 90.35 & 90.54 & \textbf{88.92} & 90.22 & 90.11 & 89.38 & 78.14 & 83.36 & 88.33 & 85.25 \\
TargetDemos + SL & \underline{85.70} &\underline{73.55} & 91.77 & \underline{92.79} & 88.55 & 91.02 & \textbf{91.58} & \textbf{90.70} & 79.10 & 85.11 & \underline{88.77} & \underline{87.15} \\
TargetDemos + Rev. & 84.50 & 72.03 & 91.00 & 92.35 & \underline{88.70} & 91.06 & \underline{91.18} & 89.91 & \underline{79.13} & \underline{85.65} & 88.46 & 86.72 \\
TargetDemos + SL + Rev. & \textbf{85.88} & \textbf{77.15} & \textbf{92.56} & \textbf{93.03} & 88.59 & \textbf{91.85} & 90.63 & \underline{89.95} & \textbf{80.44} & \textbf{86.07} & \textbf{89.03} & \textbf{87.74} \\ \midrule \midrule

 \textbf{Topic Classification} & az & cy & de & en & he & id & ro & sl & sw & te & th & avg \\ \midrule
Gold Data & \textit{84.85} & \textit{77.03} & \textit{84.45} & \textit{89.99} & \textit{83.65} & \textit{87.77} & \textit{86.53} & \textit{86.04} & \textit{76.64} & \textit{80.29} & \textit{86.94} & \textit{84.02} \\\hline
Summarized Label (SL) & 75.96 & 56.10 & 72.68 & 72.89 & 68.70 & 72.40 & 73.27 & 72.69 & 59.77 & 67.62 & 73.92 & 69.64 \\
EnglishDemos + SL & 76.60 & 60.04 & 74.69 & 78.25 & 71.56 & 77.29 & 73.25 & 75.32 & 63.23 & 66.22 & \underline{76.48} & 72.08 \\
EnglishDemos + Rev. & \underline{77.95} & 63.70 & 75.13 & \textbf{79.35} & 71.82 & \underline{78.58} & \underline{74.89} & \underline{77.53} & \textbf{66.38} & 68.00 & \textbf{77.58} & \underline{73.72} \\
TargetDemos & 77.82 & 56.84 & 72.99 & \underline{78.58} & \underline{71.93} & 78.38 & 73.91 & 77.07 & 61.02 & 71.41 & 73.71 & 72.15 \\
TargetDemos + SL & 73.17 & 57.54 & 74.68 & 78.25 & 71.16 & 77.53 & 73.60 & 77.16 & 63.27 & 70.81 & 75.79 & 72.09 \\
TargetDemos + Rev. & \textbf{78.75} & \textbf{66.23} & \textbf{75.64} & \textbf{79.35} & 71.15 & \textbf{79.77} & \textbf{75.77} & \textbf{77.80} & \underline{63.30} & \textbf{72.24} & 74.81 & \textbf{74.07} \\
TargetDemos + SL + Rev. & 76.19 & \underline{64.13} & \underline{75.32} & 76.99 & \textbf{72.07} & 77.81 & 73.69 & 76.33 & 63.02 & \underline{71.63} & 76.27 & 73.04 \\ \midrule \midrule

 \textbf{Sentiment Analysis} & az & cy & de & en & he & id & ro & sl & sw & te & th & avg \\ \midrule
Gold Data & \textit{71.69} & \textit{58.77} & \textit{65.84} & \textit{80.97} & \textit{82.23} & \textit{90.39} & \textit{90.48} & \textit{83.40} & \textit{75.07} & \textit{82.51} & \textit{80.14} & \textit{78.32} \\\hline
Summarized Label (SL) & 61.66 & 40.30 & 65.88 & 70.23 & \textbf{75.76} & 75.49 & \underline{82.83} & 50.24 & 46.45 & 66.59 & 72.44 & 64.35 \\
EnglishDemos + SL & \textbf{67.71} & 42.46 & 65.91 & \textbf{77.18} & 70.10 & 83.34 & 82.33 & 59.52 & 43.63 & 64.13 & \textbf{77.31} & 66.69 \\
EnglishDemos + Rev. & 66.33 & 41.52 & \textbf{67.85} & 67.58 & 67.52 & 80.95 & 80.96 & 59.96 & 50.82 & 66.05 & 76.04 & 65.96 \\
TargetDemos & 64.21 & \underline{48.35} & 65.09 & \underline{76.62} & 70.38 & 83.99 & 80.47 & 66.22 & \textbf{62.02} & 65.93 & 73.14 & 68.77 \\
 TargetDemos + SL & \underline{67.56} & 42.53 & 65.37 & 70.23 & \underline{75.47} & \textbf{86.99} & 79.76 & 71.64 & 48.26 & \underline{73.11} & 75.63 & \underline{68.78} \\
TargetDemos + Rev. & 67.01 & \textbf{54.41} & \underline{66.84} & \textbf{77.18} & 74.71 & 85.18 & \textbf{83.34} & \underline{72.47} & \underline{60.34} & 72.35 & 74.10 & \textbf{71.63} \\
TargetDemos + SL + Rev. & 67.19 & 37.15 & 63.65 & 65.57 & 75.46 & \underline{86.82} & 79.74 & \textbf{76.44} & 50.34 & \textbf{74.99} & \underline{76.89} & 68.57 \\ \bottomrule
\end{tabular}
}
\caption{Fine-tuning results for the gold and artificial data averaged over 4 generative LLMs on the three tasks. For each configuration we fine-tune ten downstream XLM-R models and report average F1 scores. We bold the best generation strategy in each column and underline the second best. See \ref{sec:lang_abbreviations} for language abbreviations.}\label{tab:all-results}
\end{table*}

\begin{figure*}[h!]
\centering
\begin{subfigure}{0.32\textwidth}
\centering
\includegraphics[width = \textwidth]{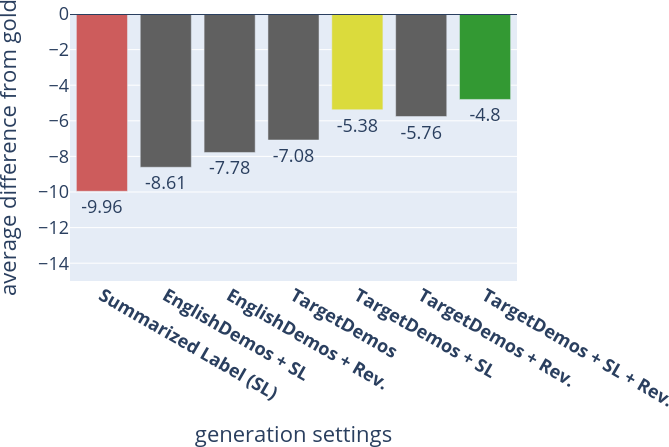}
\caption{Intent recognition.}
\label{fig:intent_task}
\end{subfigure}
\begin{subfigure}{0.32\textwidth}
\centering
\includegraphics[width = \textwidth]{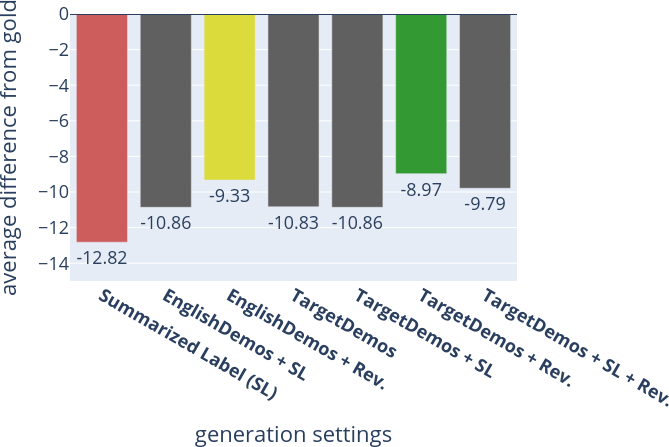}
\caption{Topic classification.}
\label{fig:topic_task}
\end{subfigure}
\begin{subfigure}{0.32\textwidth}
\centering
\includegraphics[width = \textwidth]{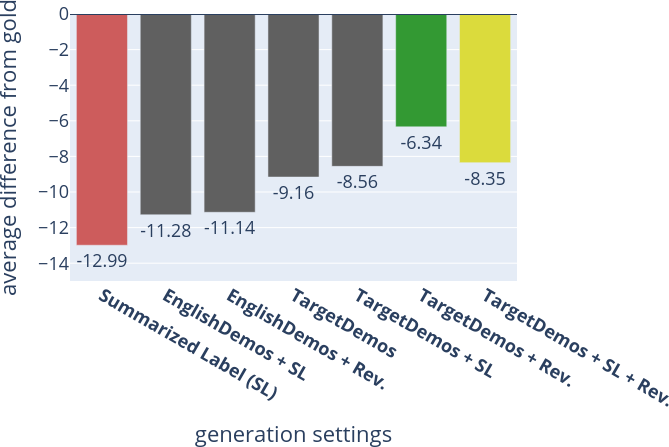}
\caption{Sentiment analysis.}
\label{fig:sentiment_task}
\end{subfigure}
\caption{Average difference in performance on the generated data compared to the gold samples across the task dimension. Smaller bars indicate better performance. Green bars show the best-performing generation setting across all languages and models for each task, and yellow bars show the second-best generation setting.}
\label{fig:combined-per-task}
\end{figure*}

\section{Results and Discussion}\label{sec:analysis}

\subsection{Overview}
Table \ref{tab:all-results} summarizes the results obtained after fine-tuning XLM-R on the LLM-generated data for each of the tasks. The first row per task indicates the scores that can be achieved after fine-tuning on the same amount of gold data, and all subsequent rows correspond to one of the generation strategies. 
Scores are averaged across four different models, see Table \ref{tab:llama3-8b-results}, \ref{tab:gemma3-4b-results}, \ref{tab:gemma3-27b-results}, and \ref{tab:llama3-70b-results} in the appendix that show more detailed downstream evaluation results per model, task, and language. 

In order to better analyze the quality of samples generated with different strategies for downstream model performance, we consider the three dimensions for a more fine-grained evaluation in the next Sections\footnote{For each dimension, we summarize results across other dimensions as a difference to the gold samples rather than absolute values, for better readability.}: Task (\ref{sec:task-dimension}), Model (\ref{sec:model-dimension}), and Language (\ref{sec:lang-dimension}) dimension.
Additionally, in Section \ref{sec:revision-strategy}, we analyse in more depth the revision strategy that proved to be most beneficial in our experiments, consistently bringing improvements across all dimensions.

\subsection{Task Dimension}\label{sec:task-dimension}

Figure \ref{fig:combined-per-task} summarizes the average difference\footnote{We list the differences as percentage points (absolute difference between F1 scores), not to be confused with percentual (relative) difference.}
in performance on three tasks for different generation strategies with the blue line on the top representing the upper bound, i.e. the fine-tuning results on the gold data. Ideally, good samples should result in very similar downstream scores. Thus, we expect the best strategy to have the smallest bar (indicated in green, while yellow is the second-best one).

\paragraph{Intent Recognition.}
For the intent recognition task (see Figure \ref{fig:intent}) \textbf{the best generation strategy includes 10 demonstrations in the target language together with a summarized label description and revision check}. XLM-R fine-tuned on the data generated in this way underperforms the model that was fine-tuned on the gold data only by 4.8\% on average. The worst generation setting for the intent classification task is when the LLM is provided only with a summarized label description without any demonstrations. However, if the summarized label description is supplemented with target demonstrations, LLMs can generate good-quality samples that result in an improvement of 4.6\% on average. Figure \ref{fig:combined-per-task} also shows that the revision check helps with both types of demonstrations: For English as well as in the target language.

\paragraph{Topic Classification.}
Using generated samples for topic classification results in a relatively large gap to the gold baseline (up to -12.8\% on average, see Figure \ref{fig:topic_task}). However, here we also observe a consistent trend that revision improves generation in all settings. \textbf{The best-performing configuration for the topic classification task includes target language demonstrations with revision}, and if demonstrations are replaced with some examples in English, the performance remains very similar ($\approx$0.4\% difference). Notably, summarized label description is detrimental to this task, which might be due to the fact that the class labels are already quite descriptive (e.g. geography, politics), and adding summarized label descriptions may introduce some noise. Besides, English demonstrations are almost as effective as demonstrations in the target language for SIB-200 which can be potentially explained by the fact that SIB-200 data were originally translated from English.

\subsubsection{Sentiment Analysis}
The sentiment analysis task represents the most diverse setting because each language has a separate dataset for evaluation that was not created as part of a unified approach like SIB-200 or MASSIVE. Although all samples are related to sentiment, texts may describe different topics, and the average length of samples differs depending on the language\footnote{For instance, gold sentiment examples in Slovenian have longer text (median value is 275 tokens) than Romanian examples (37 tokens).}. Nevertheless, this task also demonstrates the advantage of the revision strategy. Moreover, unlike the topic classification task, sentiment analysis substantially benefits from demonstrations in the target language (see Figure \ref{fig:sentiment}).

\subsection{Model Dimension}\label{sec:model-dimension}

Figure \ref{fig:combined-per-model} illustrates the performance across all tasks and languages for Gemma-3 and Llama-3 models. As expected, \textbf{larger models tend to outperform their smaller versions}, e.g. the gap between the gold baseline and the best-performing setting is -6.81\% for Llama3-8b and -5.57\% for Llama3-70b. Also, Llama models demonstrate slightly better performance across 3 tasks compared to Gemma. Overall, target language demonstrations with revision represent the best generation setting for all models, and adding only summarized label descriptions is not enough to learn useful patterns. 
Another interesting observation is that Llama benefits more from English demonstrations than Gemma. This could be attributed to the way how multilingual transformers from the Llama family construct latent representations, according to \cite{wendler-etal-2024-llamas}, their abstract ``concept space'' lies closer to English than to other languages, possibly making examples and reasoning in English more useful. 


\subsection{Language Dimension}\label{sec:lang-dimension}

The main objective of our experiments is to identify the best generation strategies for a low-resource scenario when we have very little (if any) annotated gold data. However, the low-resource status applies not only to the amount of training data available for the task but also to the target language. Models are expected to be better at generating data in high-resource languages like English or German because they were pre-trained on a large amount of data in those languages. In fact, the results of Figure \ref{fig:combined-per-language} in the Appendix indicate that English and German achieve relatively good scores in all generation settings, and their best scores are very close to the gold baseline (-3.1\% for German and -5.8\% for English on average) --  synthetic data work almost as good as original data.

\begin{figure*}[h!]
\centering
\begin{subfigure}{0.32\textwidth}
\centering
\includegraphics[width = \textwidth]{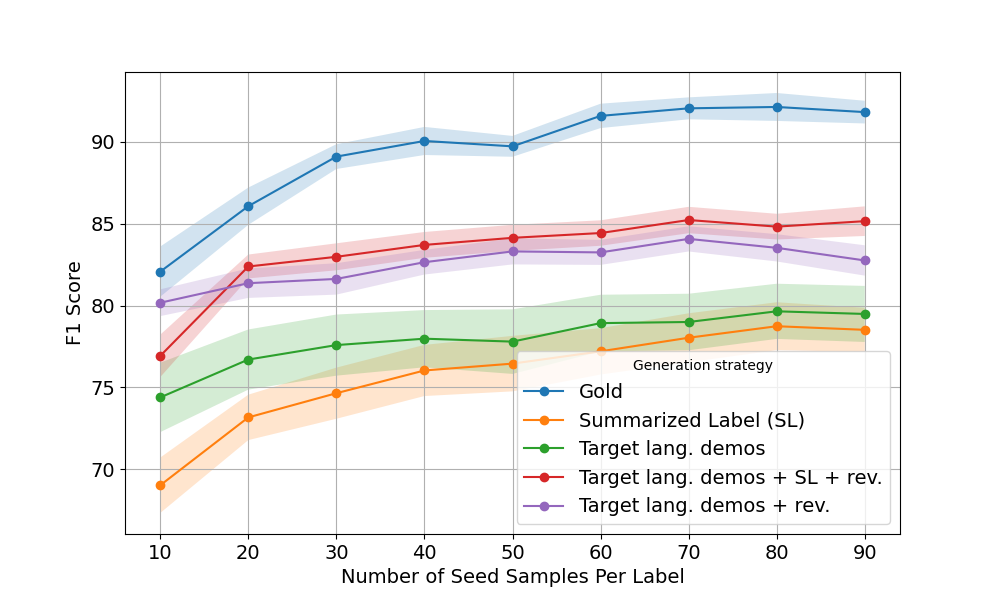}
\caption{Intent recognition Azerbaijani.}
\label{fig:per_seed_intent_az}
\end{subfigure}
\begin{subfigure}{0.32\textwidth}
\centering
\includegraphics[width = \textwidth]{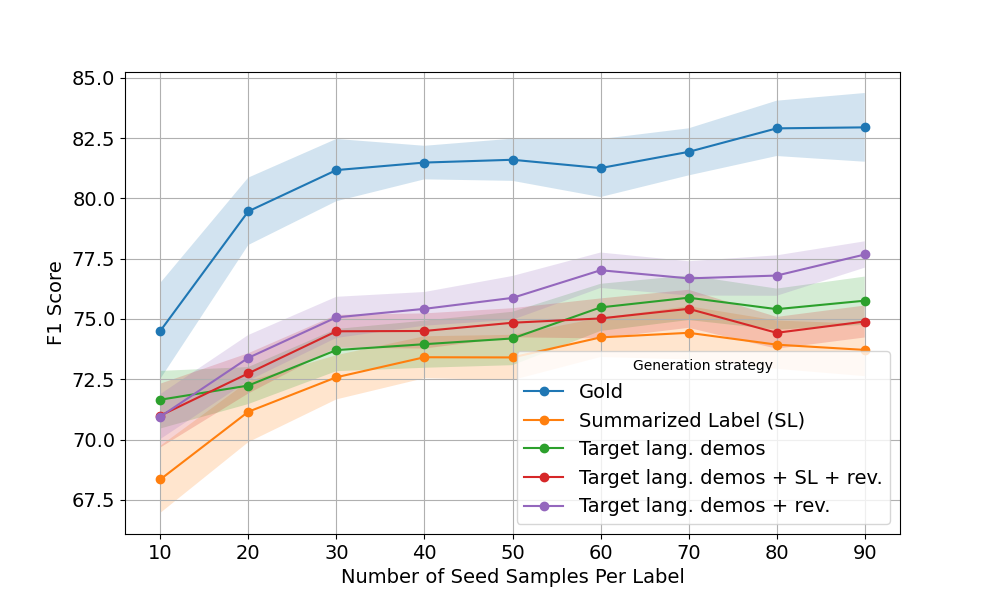}
\caption{Topic classification Azerbaijani.}
\label{fig:per_seed_topic_az}
\end{subfigure}
\begin{subfigure}{0.32\textwidth}
\centering
\includegraphics[width = \textwidth]{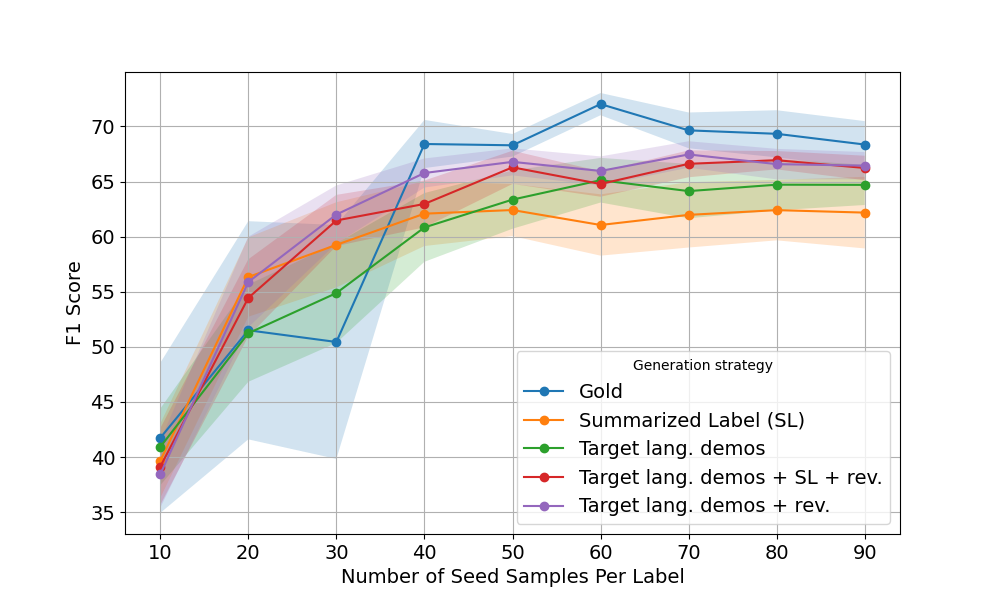}
\caption{Sentiment analysis Azerbaijani.}
\label{fig:per_seed_sent_az}
\end{subfigure}
\begin{subfigure}{0.32\textwidth}
\centering
\includegraphics[width = \textwidth]{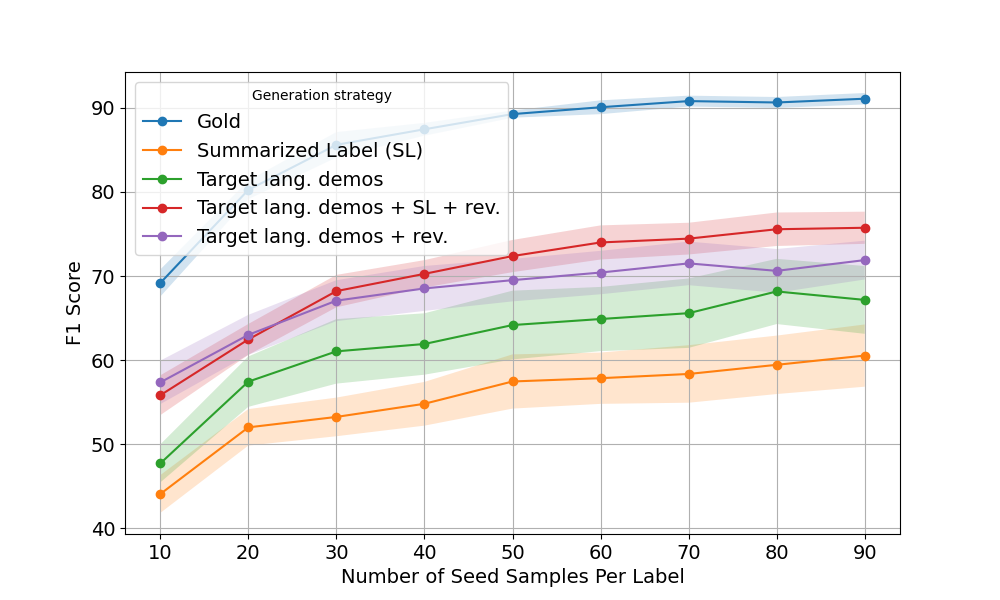}
\caption{Intent recognition Welsh.}
\label{fig:per_seed_intent_cy}
\end{subfigure}
\begin{subfigure}{0.32\textwidth}
\centering
\includegraphics[width = \textwidth]{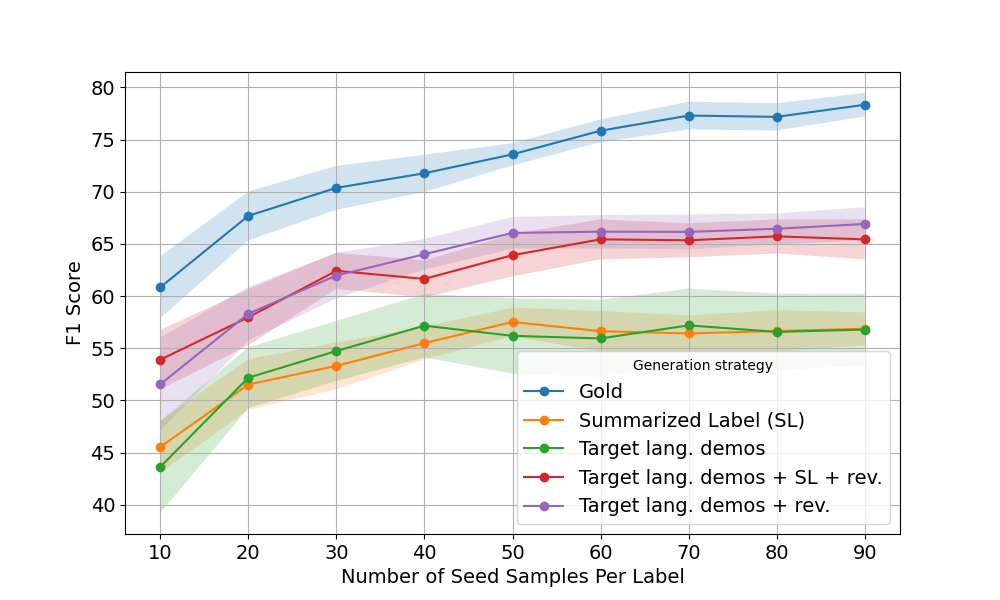}
\caption{Topic classification Welsh.}
\label{fig:per_seed_topic_cy}
\end{subfigure}
\begin{subfigure}{0.32\textwidth}
\centering
\includegraphics[width = \textwidth]{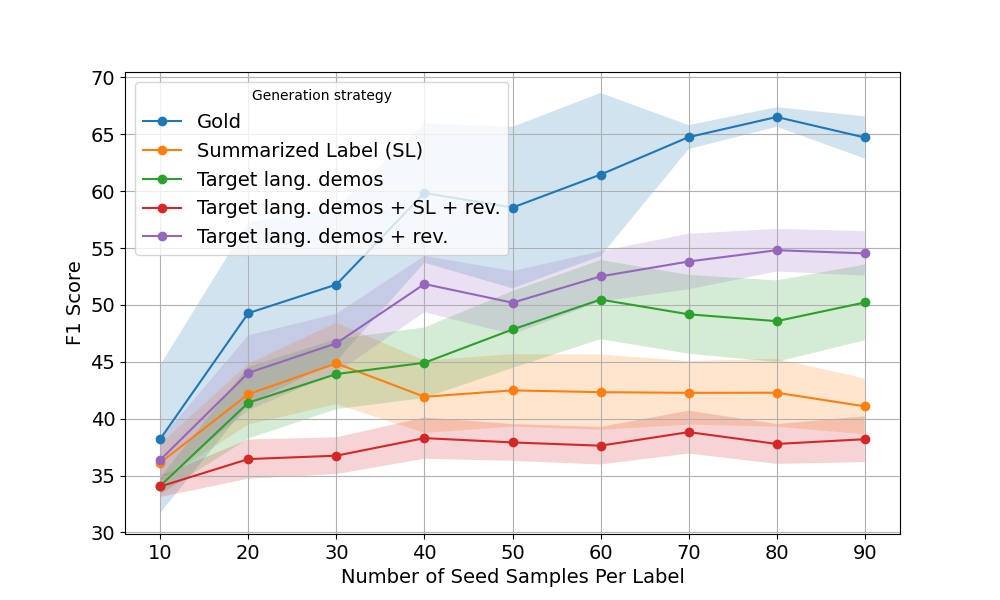}
\caption{Sentiment analysis Welsh.}
\label{fig:per_seed_sent_cy}
\end{subfigure}
\caption{F1 mean performance with 95\% Conf. Intervals for low-resource languages of Azerbaijani and Welsh for different numbers of seed samples per label used. We visualize the most prominent generation strategies and the gold data for comparison.}
\label{fig:perf_per_seed}
\end{figure*}

However, for lower-resource languages, the performance gap gets larger, e.g., for Welsh, even the best generation setting underperforms the gold baseline by -11.53\% on average. 
Slovenian, Indonesian, Hebrew, Thai, and Telugu achieve the best scores \textbf{with a combination of target language demonstrations, intent description and revision}. Azerbaijani, Welsh, Romanian, Swahili and English achieve the highest scores with target language demonstrations and revision but without any intent description. Interestingly, German is the only language that shows slightly better performance when demonstrations are in English. These findings are in line with previous work that attributes an often low performance to LLMs on extremely low-resource languages \cite{gurgurov2025small}.

English demonstrations without revision perform the worst for Swahili and Telugu, which can most likely be addressed to the lack of pretraining data of such languages. 
For certain languages (e.g. Indonesian, Thai), XLM-R achieves comparably high F1 scores. This is most likely due to the amount of data in these languages being more prominent in the pretraining corpus of XLM-R ~\cite{DBLP:journals/corr/abs-1911-02116}.

We also analyse the effect of the number of seed samples per label collected on model performance for two of the low-resource languages, Azerbaijani and Welsh in Figure~\ref{fig:perf_per_seed}. We compare the best-generation strategies with the worst (summarized label) and gold as our baseline. The generation strategies tend to stagnate in performance around 50-60 samples per label collected while adding more gold data generally increases performance.

In summary, \textbf{using target language demonstrations together with LLM-based revision yields the best results across both task and model dimensions}. This is also the case for most languages: This combination of generation strategies is best for 4 out of 11 languages and second-best for 5 out of 11 languages. Certain exceptions exist: for languages like Thai or Hebrew, this combination is not even the second-best generation strategy. Additionally, we note, that even a small amount of demonstrations given to the LLM in the target languages increases the quality of generated data for downstream model performance substantially when compared to English demonstrations. This is the case in nearly all cases. However, the inclusion of English demonstration is also generally better for data quality compared to cases where only summarized label descriptions are used.  

\begin{figure*}[h!]
\centering
\begin{subfigure}{0.49\textwidth}
\centering
\includegraphics[width = \textwidth]{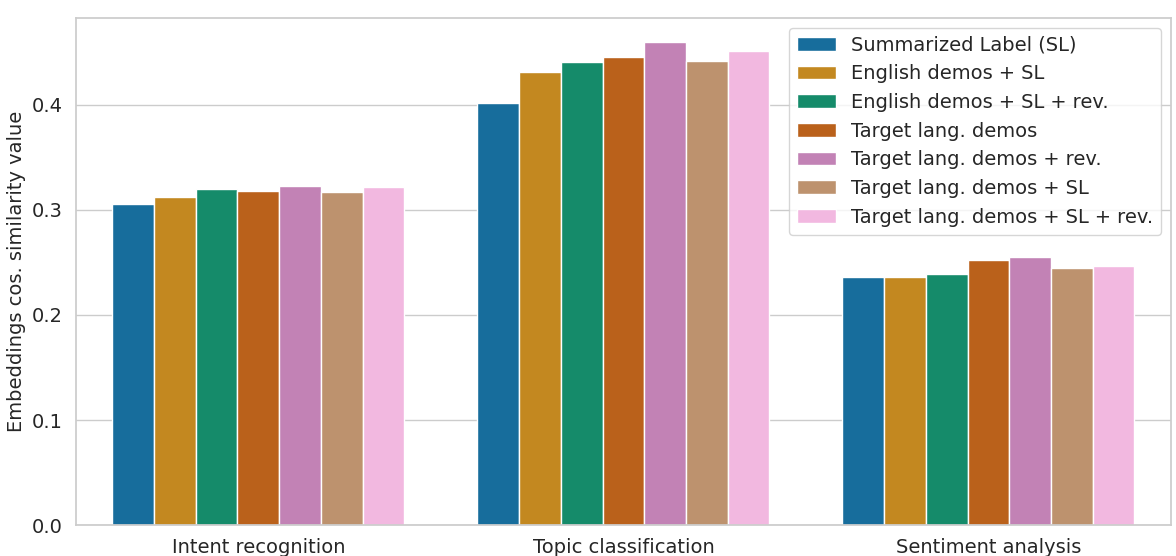}
\caption{Embedding cosine similarity to gold data.}
\label{fig:intent}
\end{subfigure}
\begin{subfigure}{0.49\textwidth}
\centering
\includegraphics[width = \textwidth]{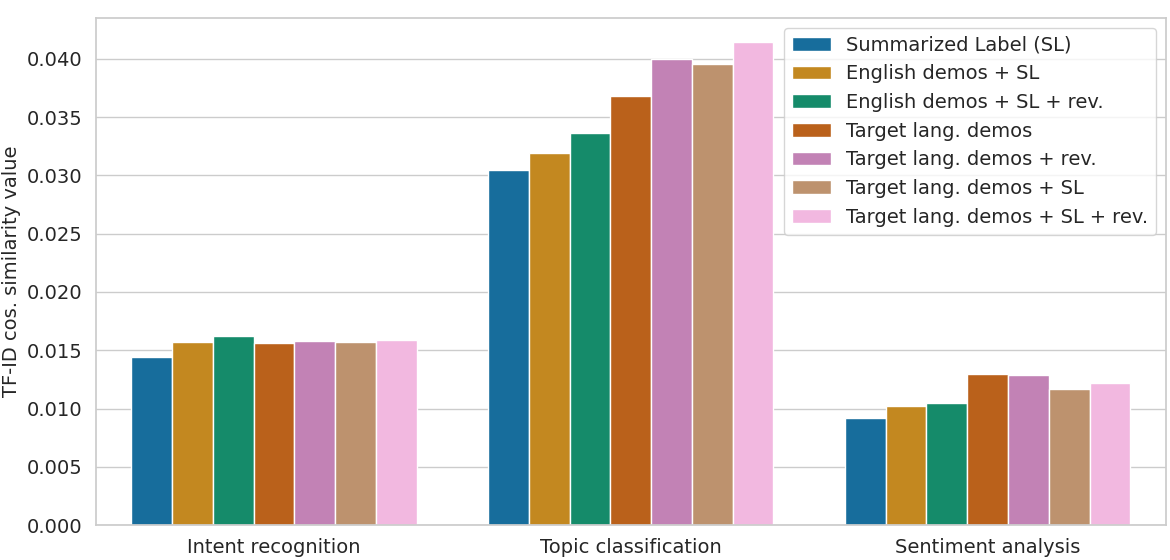}
\caption{TF-IDF cosine similarity to gold data.}
\label{fig:sentiment}
\end{subfigure}
\caption{Similarity of data generated through various generation strategies to gold data for two different similarity metrics. The figures display results averaged across languages and models.}
\label{fig:metric-similarity}
\end{figure*}

\subsection{Analysis of LLM Revision Strategy}\label{sec:revision-strategy}

Revision consistently improves generation quality as demonstrated in the downstream performance comparison across different tasks, models and languages. Table \ref{tab:num-rejected} shows how many samples are rejected in the revision cycle by each of the models. For Llama models, we can see a clear pattern that a smaller version with 8 billion parameters rejects more samples than the model with 70 billion parameters. 
For instance, for the topic classification task, Llama3-8b rejected 21\% of all generated samples while Llama3-70b rejected only 0.89\%. Similarly, for the same task, Gemma3-4b rejected 24.07\% of samples, and Gemma3-27b rejected only 9.64\%. However, for sentiment analysis, both versions of Gemma reject approximately the same amount (around 7.5\%), while Llama models retain the pattern where the larger version has a much lower rejection rate (only 1\%).

Our manual analysis revealed that the rejected samples typically describe related concepts that are thematically relevant but do not fully describe the main intent (e.g. travel expenses and booking in Figure \ref{fig:revision-example} are only indirectly related to travel). The revision also helps to filter out the cases that are ungrammatical, have incorrect output language, or are irrelevant to the intent, which is often the problem for low-resource languages. 

Based on the rejection proportions across languages (see Table \ref{tab:num-rejected}) we can conclude that generation quality differs substantially depending on the language. For example, Welsh samples have a high rate of rejection (up to 65.43\% for the topic classification task with Gemma3-4b) while English samples have comparatively low rejection rates. 

\section{Additional Evaluation: Similarity of Generated Data to Gold Data}\label{sec:similarity-to-gold}

In order to get an idea of how close the generated data is semantically to the original data, we measure the average similarity of each generation strategy to gold data in Figure~\ref{fig:metric-similarity}, where we distinguish between various tasks in our study. We compute two metrics: The cosine similarity of TF-IDF~\cite{Rajaraman_Ullman_2011} and the cosine similarity of embeddings. For the embedding computation, we used a multilingual embedding model~\cite{feng-etal-2022-language} and calculated similarity pairwise, using the samples with the same label, and then averaged the scores.

Notably, LLM revision increases the similarity of the generated data with regard to gold data in nearly all cases where revision is used. This is most notable for topic classification, while the difference in similarity for intent recognition and sentiment analysis is quite low. Target language demonstrations also lead to generally higher similarity between the gold and the generated data.

Interestingly, although the similarity to the gold data between the different generation strategies for the intent recognition task does not vary much, some strategies still result in better performance than others, with target demonstrations and revision being the most beneficial, as demonstrated in Figure \ref{fig:intent_task}. This indicates that, depending on the task, generated examples can be useful for the downstream task even when they are not very similar to the original gold samples, potentially even adding more diversity to the training data.

We note that the Pearson correlation coefficient between the F1 performance of individual methods and their cosine similarity to human data correlates positively, but only very weakly (0.111 on average across tasks and languages).

To check our hypothesis that generated data add more diversity, we evaluated n-gram diversity of the gold and generated training samples and found that the diversity of the data generated with revision and demonstrations is on average +0.218 higher for the intent recognition task, and it is also higher for the generated sentiment analysis data (+0.581), although the metric is slightly lower for the topic classification (-0.279 compared to the gold data). This can be explained by the fact that the sentiment data includes the most diverse samples that were sourced from different datasets, varying in length and topic coverage. Also, data generated with Gemma demonstrates, on average, better diversity than Llama-generated data, especially with the larger version of Gemma that has 27B parameters.

\section{Conclusion}

In this work, we perform a comparative analysis of seven data generation strategies that combine summarised label descriptions, language of demonstrations, and LLM-based revision. We benchmark the models fine-tuned on the data generated with these strategies on three NLP tasks and eleven languages, including several low-resourced ones. Our experiments reveal that the best strategy, in general, is the usage of target language demonstrations with revision: revision consistently improves the quality of generated samples even in the low-resource scenario. Next, for the Llama models, target language demonstrations can be replaced with examples in English for comparable performance. We also found that while intent recognition tasks benefit from the inclusion of label descriptions, generating the data for topic classification works better without them.

As expected, larger models tend to generate better quality examples, and Llama models slightly outperform Gemma. However, even though data generated for higher-resource languages has better quality, using an optimal generation strategy may substantially reduce the performance gap. 
Our findings highlight the importance of careful selection of the data generation strategy, as we provide a recipe for creating generation prompts in a low-resource setting for different tasks and languages.


\section*{Limitations}
Due to the scope of this study and the large number of tested configurations, we experiment with only two model families with good multilingual capabilities: Gemma-3 and Llama-3. It is possible that some other models, e.g. Qwen-3, will perform better in the generation setting, at least for some of the languages. We also limited the study to only 11 languages due to resource constraints and the availability of evaluation data for the low-resource languages. 

We could also consider all available labels in the MASSIVE dataset (currently subsampled to 10). However, we believe that such simplifications were necessary to avoid many conflating factors that can be potentially caused by the semantic overlaps in label descriptions.

We additionally did not consider using English demonstrations without revision. Target language demonstrations with revision consistently outperform English demonstrations with revision, and as such, we would expect similar performance for English demonstrations without revision compared to target language demonstrations without revision, making their usage in this study redundant.

Another limitation of this work is the lack of human annotations for checking the quality of the generated data. Involving human annotators was not feasible in the scope of the current project, given the amount of generated data for multiple languages, tasks, models, and generation configurations. Besides, many of the languages included in our study are very low-resourced (e.g. Welsh, Azerbaijani, Telugu), and it would be difficult to find native speakers to perform human evaluation for them. We consider a follow-up study with human annotators as future work, and in the current project, we opt for an automated evaluation using the downstream performance on the gold test data (that was human-annotated) as a performance indicator.

We recognise that the extent of potential data contamination is currently unknown because the data on which LLMs were trained is not fully disclosed, which might be reflected in the disparities between the LLMs' performance when generating data in different languages and domains. We believe that measuring this reliably is out-of-scope for our study, but we recognise this as a potential problem.


\section*{Acknowledgments}
This work was supported by the Ministry of Education, Youth and Sports of the Czech Republic through the e-INFRA CZ (ID:90254); by the German Federal Ministry of Research, Technology and Space (BMFTR) as part of the project TRAILS (01IW24005); by DisAI and AI-CODE, projects funded by European Union under the Horizon Europe, GA No. 101079164 and No. 101135437; by the European Union NextGenerationEU through the Recovery and Resilience Plan for Slovakia under the projects No. 09I01-03-V04-00059.
and No. 09I03-03-V03-00020.
\bibliography{custom}

\clearpage
\appendix

\section{Appendix}

\subsection{Ethical Considerations}
Based on a thorough ethical assessment performed on the basis of intra-institutional ethical guidelines and checklists tailored to the use of data and algorithms, we see no ethical concerns pertaining directly to the conduct of this research. Although the production of new data through LLMs bears several risks, such as the introduction of biases, the small size of the produced dataset, sufficient for experimentation, is, at the same time, insufficient for any major machine learning endeavours where such biases could be transferred.

We follow the license terms for all the models and datasets we used (such as the one required for the use of the Llama-3 model) – all models and datasets allow their use as part of the research.

\subsection{Language Abbreviations}\label{sec:lang_abbreviations}
\begin{table}[h]
\centering
\footnotesize
\begin{tabular}{ll}
\toprule
Code &  Language \\
\midrule
    az & Azerbaijani \\
    cy & Welsh \\
    de & German \\
    en & English \\
    he & Hebrew \\
    id & Indonesian \\
    ro & Romanian \\
    sl & Slovenian \\
    sw & Swahili \\
    te & Telugu \\
    th & Thai \\
\bottomrule
\end{tabular}
\caption{Language abbreviations.}
\label{tab:lang_abbreviations}
\end{table}


\subsection{Computational Resources}\label{sec:computational_resources}
All generation experiments were done using vllm \cite{kwon2023efficient} for efficient inference. For a single language-model combination it takes approximately 2 hours to generate the data for intent recognition, 1 hour 20 minutes for topic classification and 23 minutes for sentiment classification. We use a single H100 GPU to generate the samples. Thus, for 11 languages, 4 models and 3 different tasks it takes around 157 hours (6.5 days) to generate all the data.

The fine-tuning experiments were performed on RTXA6000 GPU, and XLM-R fine-tuning for each setting with 10 different seeds takes around 120 minutes for intent recognition, 95 minutes for topic classification and 30 minutes for sentiment analysis. Note that all computations can be done in parallel. 

\subsection{Intent Recognition Task: Examples}\label{app:massive-examples}

\begin{table}[h]
    \centering
    \footnotesize
    \begin{tabular}{l|p{3.8cm}}
        \textbf{label} & \textbf{example} \\\hline
        datetime\_convert & \textit{tell me the time in g. m. t. plus five}\\\hline
        alarm\_query & \textit{please list active alarms}\\\hline
        weather\_query & \textit{how's the weather like in beijing}\\\hline
        audio\_volume\_down & \textit{you're too loud}\\\hline
        play\_audiobook & \textit{i want to start war and peace where i left off}\\\hline
        cooking\_recipe & \textit{what's needed to make pizza} \\\hline
        recommendation\_movies & \textit{name a rom com movie playing in and around new york theaters} \\\hline
        transport\_ticket & \textit{olly i need to get to bristol friday night can you book me a ticket please} \\\hline
        email\_sendemail & \textit{could you please gather a list of local restaurants and email them to my husband} \\\hline
        calendar\_remove & \textit{remove stand-up on friday at ten am}
    \end{tabular}
    \caption{Labels from the MASSIVE dataset.}
    \label{tab:massive-labels}
\end{table}


\subsection{Topic Classification Task: Examples}\label{app:sib200-examples}

\begin{table}[h]
    \centering
    \footnotesize
    \begin{tabular}{l|p{4.5cm}}
    \textbf{label} & \textbf{example} \\\hline
     health & \textit{Danielle Lantagne, a UN expert on the disease, stated the outbreak was likely caused by the peacekeepers.}\\\hline
     geography & \textit{Large areas further north are quite sparsely populated, and some are nearly uninhabited wilderness.}\\\hline
     politics & \textit{Soon after the outbreak of hostilities, Britain initiated a naval blockade of Germany.}\\\hline
     entertainment & \textit{``She's very cute and sings quite well, too,'' he said according to a transcript of the news conference.}\\\hline
     science/technology & \textit{The atom can be considered to be one of the fundamental building blocks of all matter.}\\\hline
     sports & \textit{Massa is due to be out for at least the rest of the 2009 season.}\\\hline
     travel & \textit{Over 60 cruise ships ply the Galapagos waters - ranging in size from 8 to 100 passengers.}\\
    \end{tabular}
    \caption{Labels from the SIB-200 dataset.}
    \label{tab:sib200-labels}
\end{table}


\subsection{Downstream Fine-tuning of XLM-R}\label{sec:downstream_finetuning}
For the downstream evaluation we fine-tune XLM-R \cite{DBLP:journals/corr/abs-1911-02116} \textit{FacebookAI/xlm-roberta-base} model for 50 epochs with a batch size 16 and employ early stopping with a patience of 5 epochs to prevent overfitting. We perform hyperparameter optimization to determine the optimal learning rate and set it to 1e-5, \textit{AdamW} is used as an optimizer. Our dataset is balanced in both gold and generated settings and includes 100 samples per label. We normalize all inputs by converting them to lowercase and removing punctuation. We fine-tune ten models per generation strategy for all language, task and model combinations, and report average F1 scores.


\subsection{Summarized Label Examples}\label{sec:summarized_intent_examples}

This section shows the examples of summarized label descriptions (generated with Llama3-70b) for each task. Table \ref{tab:summarized_intent_examples_massive} demonstrates generated descriptions for the intent recognition task, Table \ref{tab:summarized_intent_examples_sib200} for topic classification, and Table \ref{tab:summarized_intent_examples_sentiment} for sentiment analysis.

\begin{table*}[]
\centering
\footnotesize
\begin{tabular}{l|p{10cm}}
\textbf{intent}        & \textbf{description} \\\hline
alarm\_query           & This intent involves querying or checking the status of existing alarms, including their timing, scheduling, and active status, to stay informed and manage one's alarm settings.                                   \\\hline
audio\_volume\_down    & This intent involves requesting to reduce or lower the volume of an audio device, such as a speaker, to a softer level.                                                                                             \\\hline
calendar\_remove       & This intent involves requesting to delete, remove, or cancel a specific event or all events from a calendar.                                                                                                        \\\hline
cooking\_recipe        & This intent involves seeking instructions or information about preparing and cooking various dishes, including specific ingredients, cooking times, temperatures, and methods.                                      \\\hline
datetime\_convert      & This intent involves converting or comparing time zones, including calculating time differences between specific locations or zones, or adjusting a given time to a different time zone.                            \\\hline
email\_sendemail       & This intent involves requesting to send an email to a specific recipient, often with a mention of the content or purpose of the email, such as sharing information, making a request, or inquiring about something. \\\hline
play\_audiobook        & This intent involves requesting to play, resume, or restart an audiobook, often specifying the title or author, and may include additional actions like going back to a previous point in the audiobook.            \\\hline
recommendation\_movies & This intent involves asking for movie recommendations or inquiring about current movie showtimes and ratings, often with a focus on the user's location or specific preferences.                                    \\\hline
transport\_ticket      & This intent involves booking, purchasing, or searching for train tickets, often specifying details such as destination, travel date, and payment method.                                                            \\\hline
weather\_query         & This intent involves asking about the current or future weather conditions, including temperature, precipitation, and other weather-related information, often specifying a particular location or date.         
\end{tabular}
\caption{Summarized label (intent) examples for the intent recognition task.}
\label{tab:summarized_intent_examples_massive}
\end{table*}


\begin{table*}[]
\centering
\footnotesize
\begin{tabular}{l|p{10cm}}
\textbf{intent}        & \textbf{description} \\\hline
entertainment      & This intent involves discussing leisure activities, hobbies, and forms of entertainment, such as boating, music, comedy, and travel, as well as mentioning celebrities, events, and notable achievements in the entertainment industry. \\\hline
geography          & This intent involves describing or providing information about geographical locations, features, and formations, including cities, islands, mountains, and landforms, as well as their characteristics, sizes, and relationships to each other. \\\hline
health             & This intent involves discussing various aspects of health, including infectious and contagious diseases, their causes, symptoms, and prevention methods, as well as medical treatments, diagnostic tools, and personal health care practices. \\\hline
politics           & This intent involves discussions, statements, and actions related to government policies, political leaders, elections, and legislation, often referencing specific events, decisions, or opinions that shape or are shaped by political institutions and processes. \\\hline
science/technology & This intent involves discussing or explaining scientific concepts, theories, and discoveries in various fields such as astronomy, medicine, biology, and technology, often providing historical context and details of experiments or research that support these concepts. \\\hline
sports             & This intent involves discussing various sports and athletic activities, including running, hockey, cross-country, skiing, fencing, golf, and others, as well as mentioning specific athletes, teams, and competitions. \\\hline
travel             & This intent involves planning, preparing for, or experiencing various modes of transportation, trips, and travel-related activities, including road travel, air travel, boat trips, and cruises, as well as considering factors that may impact travel, such as weather, road conditions, and cultural differences.
\end{tabular}
\caption{Summarized label (intent) examples for the topic classification task.}
\label{tab:summarized_intent_examples_sib200}
\end{table*}

\begin{table*}[]
\centering
\footnotesize
\begin{tabular}{l|p{10cm}}
\textbf{intent}        & \textbf{description} \\\hline
negative & This intent involves expressing negative sentiments, complaints, or criticisms about various topics, including work, relationships, social issues, and personal struggles. The examples convey frustration, disappointment, and even anger towards certain situations or individuals. \\\hline
positive & This intent involves expressing positive sentiments, gratitude, and appreciation for various services, events, and announcements. The intent is to share good news, express joy, and acknowledge the efforts of individuals or organizations, often with a tone of optimism and enthusiasm.
\end{tabular}
\caption{Summarized label (intent) examples for the sentiment classification task.}
\label{tab:summarized_intent_examples_sentiment}
\end{table*}


\subsection{Prompt Templates}

\begin{figure*}
    \centering
    \includegraphics[width=0.7\linewidth]{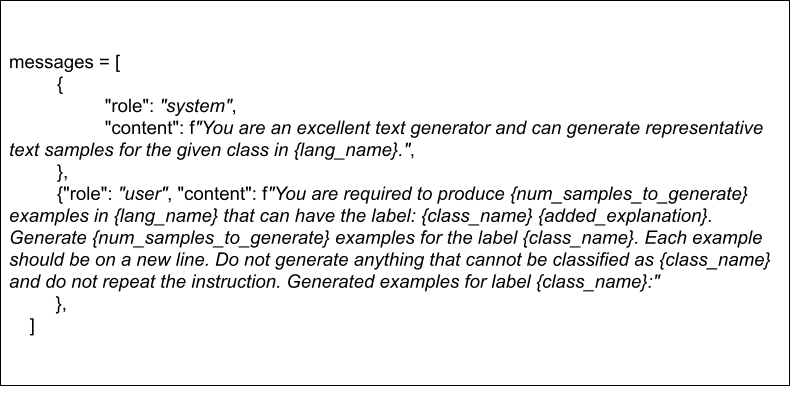}
    \caption{Generation prompt.}
    \label{fig:generation-prompt}
\end{figure*}

\begin{figure*}
    \centering
    \includegraphics[width=0.7\linewidth]{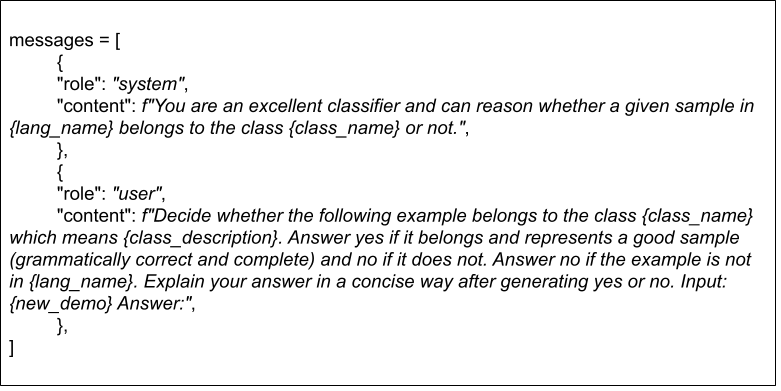}
    \caption{Revision prompt.} 
    \label{fig:revision-prompt}
\end{figure*}

\begin{figure*}
    \centering
    \includegraphics[width=0.7\linewidth]{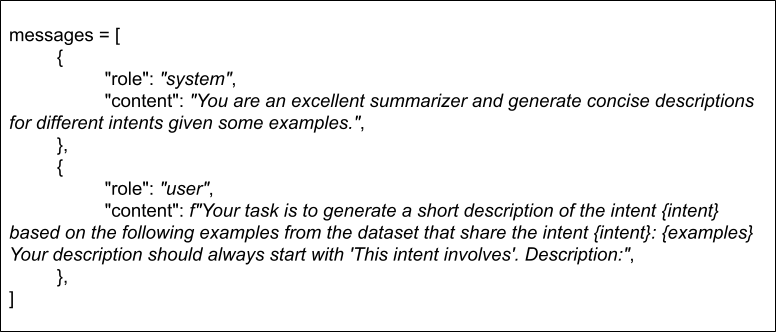}
    \caption{Summarized label generation prompt.}
    \label{fig:summarized_label_template}
\end{figure*}

This section provides the prompt templates for sample generation (Figure \ref{fig:generation-prompt}), revision (Figure \ref{fig:revision-prompt}), and summarized label generation (Figure \ref{fig:summarized_label_template}).

\subsection{Samples Rejected by Revision}\label{sec:rejected_samples}
Table \ref{tab:num-rejected} demonstrates the proportion of rejected samples in the revision setting.

\begin{table*}[h]
\resizebox{\textwidth}{!}{
\footnotesize
\begin{tabular}{@{}l|l|l|l|l|l|l|l|l|l|l|l|l@{}}
\toprule
            & \multicolumn{4}{|c|}{Topic Classification}        & \multicolumn{4}{|c|}{Sentiment Analysis}          & \multicolumn{4}{|c}{Intent Recognition}          \\ \midrule
            & Gemma3-4b & Gemma3-27b & Llama3-8b & Llama3-70b & Gemma3-4b & Gemma3-27b & Llama3-8b & Llama3-70b & Gemma3-4b & Gemma3-27b & Llama3-8b & Llama3-70b \\\hline
Azerbaijani & 34.29     & 12.86      & 40.71     & 1.14       & 14.50      & 9.50        & 46.50      & 1.50       & 22.50     & 28.80      & 43.60     & 4.30        \\
Welsh       & 65.43     & 37.29      & 55.43     & 2.29       & 53.00        & 55.50      & 55.00     & 0.00       & 65.30     & 66.00      & 62.10     & 2.10        \\
German      & 17.29     & 8.86       & 11.57     & 0.71       & 0.50       & 1.50       & 3.50      & 0.50       & 9.20      & 12.80      & 22.20     & 2.70        \\
English     & 13.14     & 3.29       & 4.43      & 0.57       & 0.00      & 0.00       & 0.00      & 0.00       & 10.20     & 22.00      & 26.10     & 8.90        \\
Hebrew      & 21.00     & 3.14       & 12.00     & 0.00       & 1.00      & 0.00       & 24.00     & 0.50       & 8.20      & 8.10       & 22.30     & 0.70        \\
Indonesian  & 15.57     & 3.86       & 14.29     & 1.29       & 0.00      & 2.00       & 6.50      & 0.50       & 11.70     & 17.30      & 24.10     & 3.70        \\
Romanian    & 15.71     & 2.86       & 9.71      & 0.71       & 0.00      & 3.00       & 8.50      & 0.50       & 6.80      & 17.40      & 22.00     & 2.80        \\
Slovenian   & 15.86     & 6.29       & 16.86     & 0.57       & 0.00      & 2.50        & 26.50     & 3.50       & 21.30     & 21.20      & 20.00     & 4.10        \\
Swahili     & 29.71     & 13.86      & 32.86     & 1.71       & 9.00      & 4.00       & 50.50     & 1.50       & 40.60     & 29.40      & 43.60     & 2.90        \\
Telugu      & 17.43     & 11.00      & 9.43      & 0.14       & 1.50       & 5.00       & 11.00     & 2.00       & 8.90      & 13.50      & 2.90      & 0.30        \\
Thai        & 19.29     & 2.71       & 24.00        & 0.71       & 1.00      & 0.00       & 20.00     & 1.00       & 11.20      & 9.80       & 31.50     & 2.20        \\\hline
total \%    & \textbf{24.07}     & 9.64       & 21.03     & 0.89       & 7.32      & 7.55       & \textbf{22.91}     & 1.05       & 19.63     & 22.39      & \textbf{29.13}     & 3.15       \\ \bottomrule
\end{tabular}
}
\caption{Proportion of rejected samples in the revision generation setting with the intent description and target language demonstrations. Numbers in bold indicate the highest rejection rate per task.}\label{tab:num-rejected}
\end{table*}


\subsection{Model Dimension and Generation Settings}
\label{app:model-dimension}
\begin{figure*}[h!]
\centering
\begin{subfigure}{0.49\textwidth}
\centering
\includegraphics[width = 0.8\textwidth]{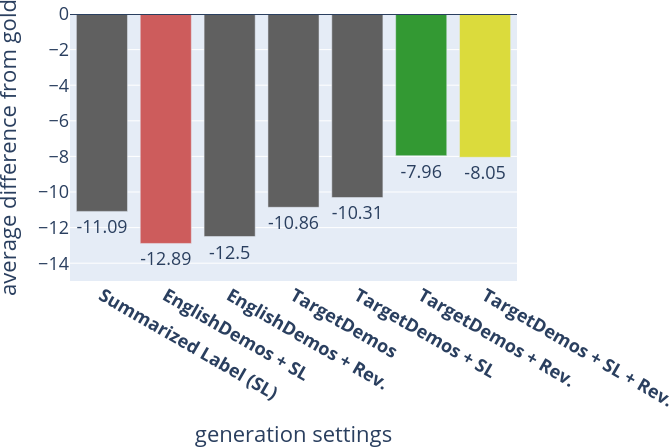}
\caption{Gemma3-4b.}
\label{fig:gemma3-4b}
\vspace{1em}
\end{subfigure}
\begin{subfigure}{0.49\textwidth}
\centering
\includegraphics[width = 0.8\textwidth]{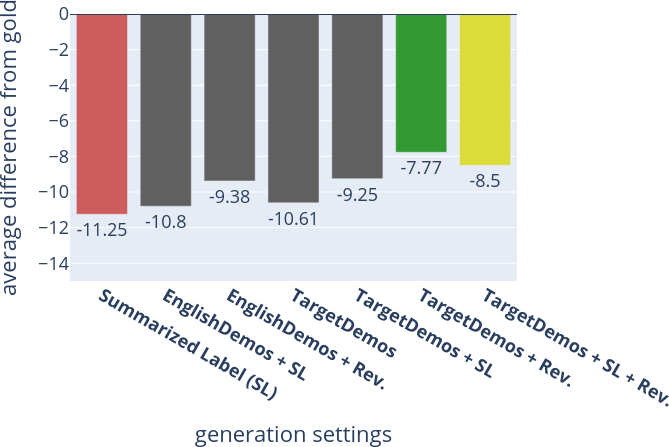}
\caption{Gemma3-27b.}
\label{fig:gemma3-27b}
\vspace{1em}
\end{subfigure}
\begin{subfigure}{0.49\textwidth}
\centering
\includegraphics[width = 0.8\textwidth]{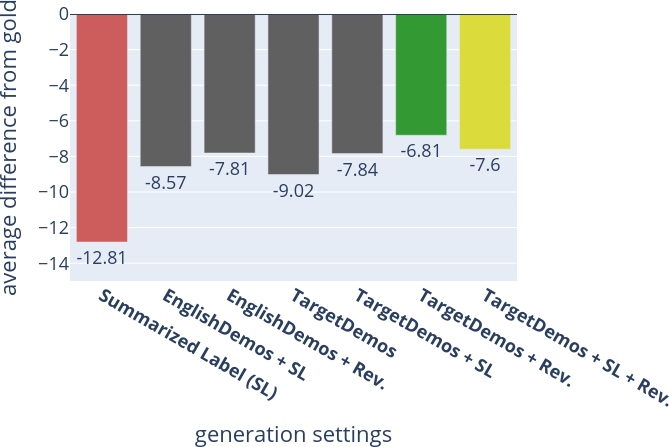}
\caption{Llama3-8b.}
\label{fig:llama3-8b}
\vspace{1em}
\end{subfigure}
\begin{subfigure}{0.49\textwidth}
\centering
\includegraphics[width = 0.8\textwidth]{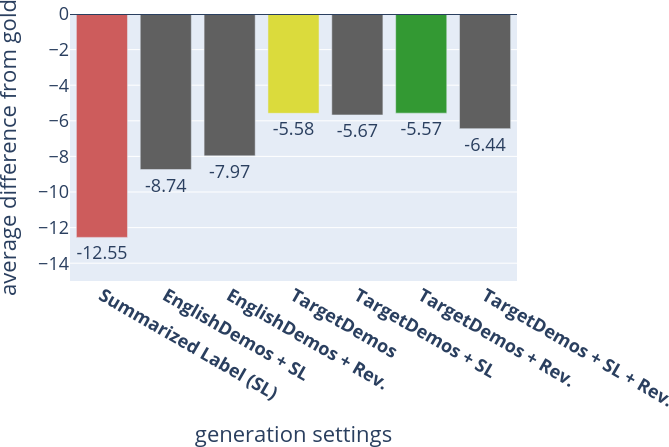}
\caption{Llama3-70b.}
\label{fig:llama3-70b}
\vspace{1em}
\end{subfigure}
\caption{Average difference in performance on the generated data compared to the gold samples across the model dimension. Lower bars indicate better performance. Green bars show the best-performing generation setting across all languages and tasks for each model, and yellow bars show the second-best generation setting.}
\label{fig:combined-per-model}
\end{figure*}

Figure \ref{fig:combined-per-model} illustrates the average difference in performance when fine-tuning the downstream model on the generated data vs. gold data fine-tuning. The results are aggregated across all languages and tasks.

\subsection{Language Dimension and Generation Settings}
\label{app:language-dimension}
\begin{figure*}
\centering
\begin{subfigure}{0.3\textwidth}
\centering
\includegraphics[width = \textwidth]{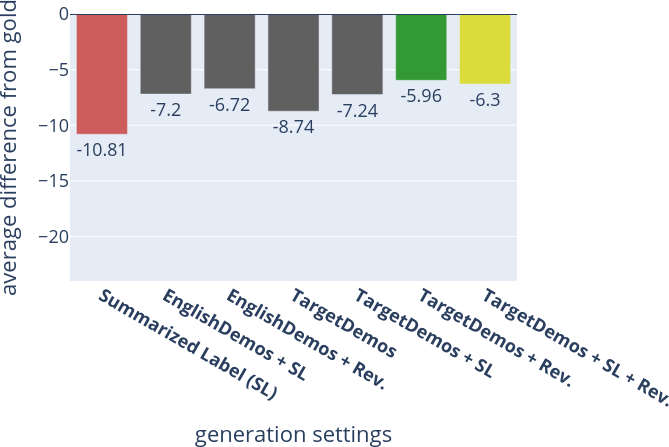}
\caption{Azerbaijani.}
\label{fig:az}
\vspace{1em}
\end{subfigure}
\begin{subfigure}{0.3\textwidth}
\centering
\includegraphics[width = \textwidth]{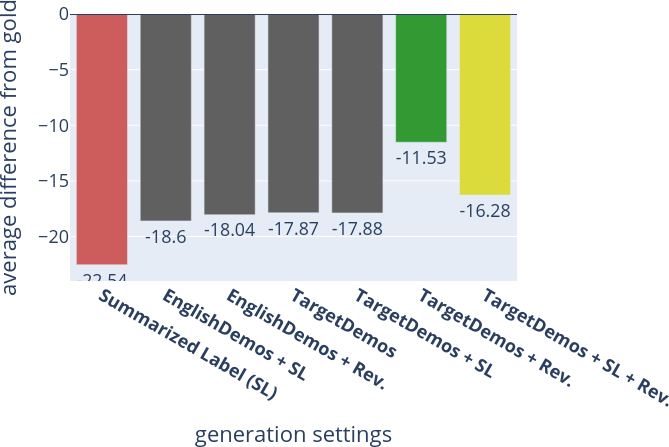}
\caption{Welsh}
\label{fig:cy}
\vspace{1em}
\end{subfigure}
\begin{subfigure}{0.3\textwidth}
\centering
\includegraphics[width = \textwidth]{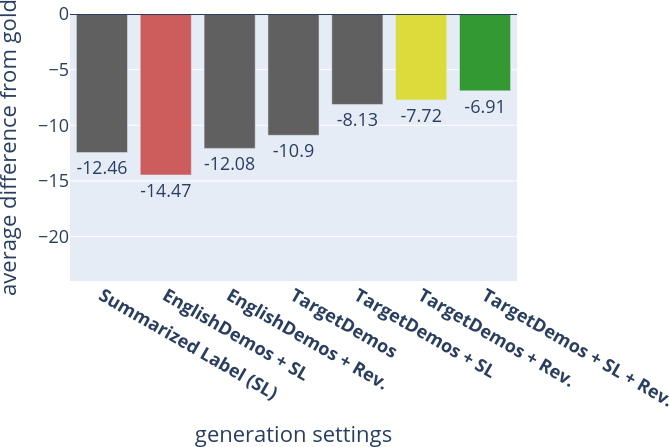}
\caption{Telugu.}
\label{fig:te}
\vspace{1em}
\end{subfigure}

\begin{subfigure}{0.3\textwidth}
\centering
\includegraphics[width = \textwidth]{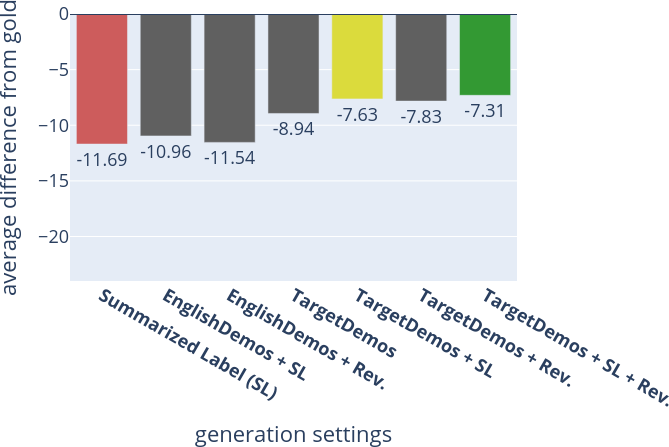}
\caption{Hebrew.}
\label{fig:he}
\vspace{1em}
\end{subfigure}
\begin{subfigure}{0.3\textwidth}
\centering
\includegraphics[width = \textwidth]{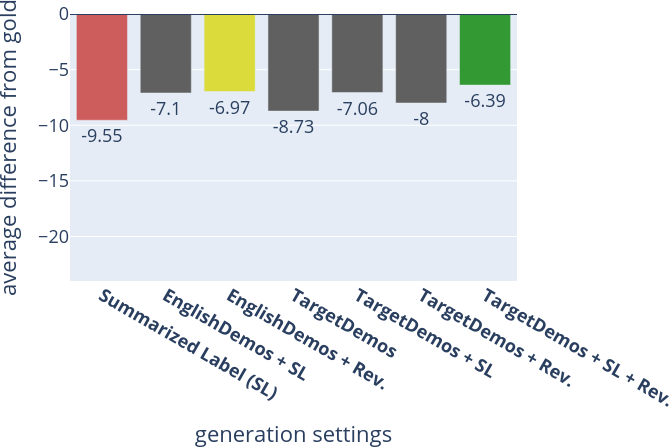}
\caption{Thai.}
\label{fig:th}
\vspace{1em}
\end{subfigure}
\begin{subfigure}{0.3\textwidth}
\centering
\includegraphics[width = \textwidth]{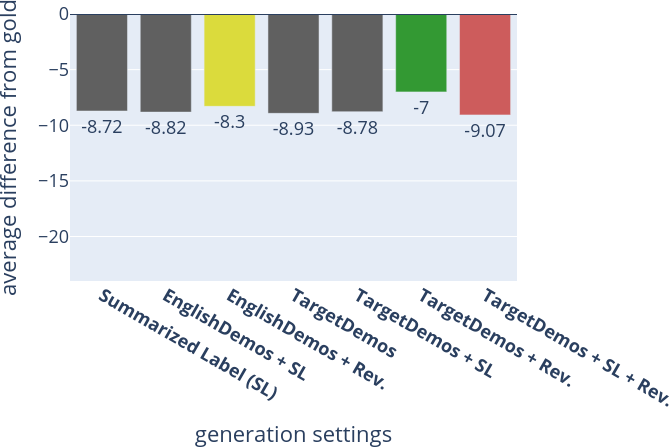}
\caption{Romanian}
\label{fig:ro}
\vspace{1em}
\end{subfigure}

\begin{subfigure}{0.3\textwidth}
\centering
\includegraphics[width = \textwidth]{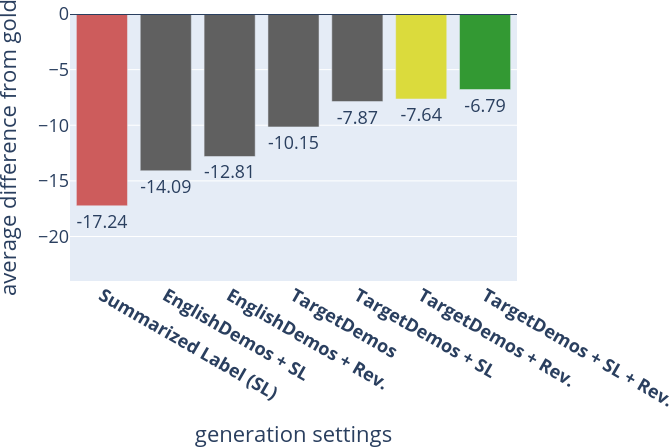}
\caption{Slovenian}
\label{fig:sl}
\vspace{1em}
\end{subfigure}
\begin{subfigure}{0.3\textwidth}
\centering
\includegraphics[width = \textwidth]{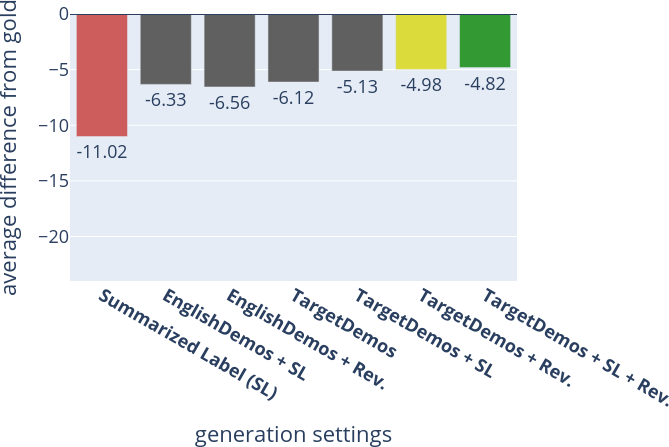}
\caption{Indonesian}
\label{fig:id}
\vspace{1em}
\end{subfigure}
\begin{subfigure}{0.3\textwidth}
\centering
\includegraphics[width = \textwidth]{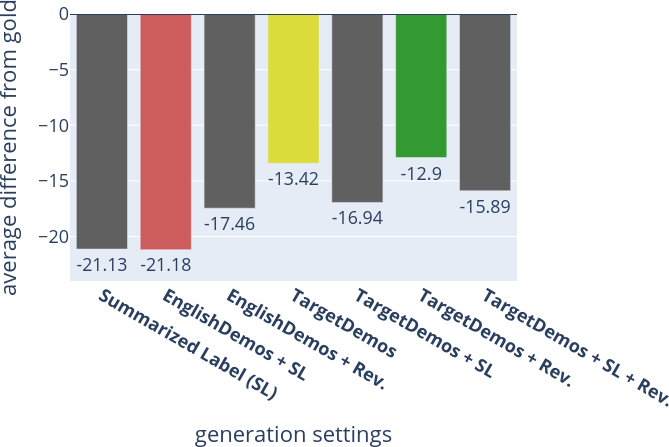}
\caption{Swahili}
\label{fig:sw}
\vspace{1em}
\end{subfigure}

\begin{subfigure}{0.3\textwidth}
\centering
\includegraphics[width = \textwidth]{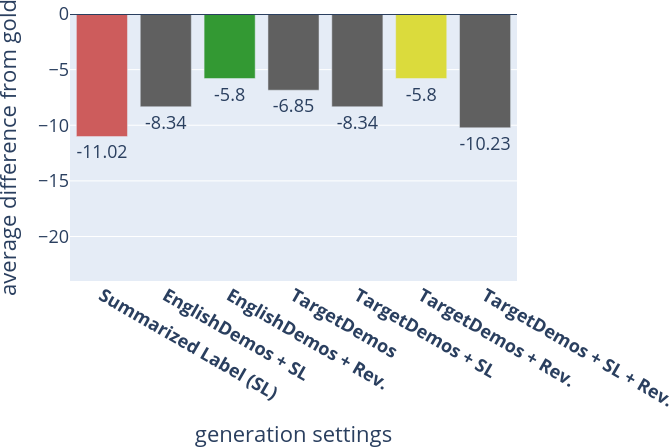}
\caption{English}
\label{fig:en}
\vspace{1em}
\end{subfigure}
\begin{subfigure}{0.3\textwidth}
\centering
\includegraphics[width = \textwidth]{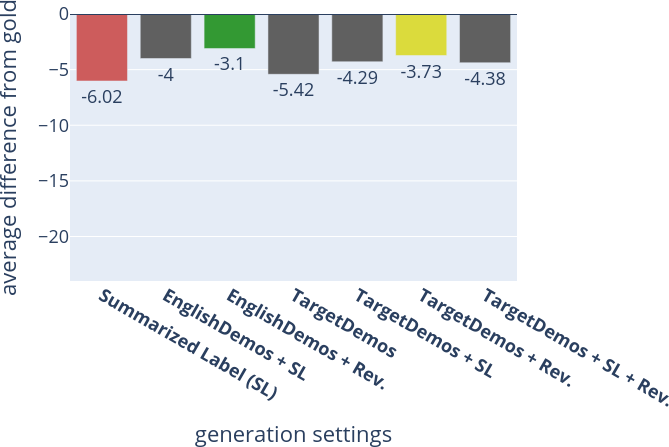}
\caption{German}
\label{fig:de}
\vspace{1em}
\end{subfigure}
\caption{Average difference in performance on the generated data compared to the gold (original) samples across the language dimension. Lower bars indicate better performance. Green bars shows the best-performing generation setting across all models and tasks for each language, and yellow bars show the second best generation setting.}
\label{fig:combined-per-language}
\end{figure*}

Figure \ref{fig:combined-per-language} illustrates the average difference in performance when fine-tuning the downstream model on the generated data vs. gold data fine-tuning. The results are aggregated across all models and tasks.

\subsection{Fine-tuning Results per Task, Model, and Language}

This section demonstrates the fine-tuning results for all tasks and languages for Llama3-8b in Table \ref{tab:llama3-8b-results}, Gemma3-4b in Table \ref{tab:gemma3-4b-results}, Gemma3-27b in Table \ref{tab:gemma3-27b-results}, and Llama3-70b in Table \ref{tab:llama3-70b-results}.

\begin{table*}[h!]
\resizebox{\textwidth}{!}{
\begin{tabular}{@{}llllllllllll|l@{}}
\toprule
\textbf{Intent Recognition} & az             & cy             & de             & en             & he             & id             & ro             & sl             & sw             & te             & th             & avg         \\ \midrule
Gold Data & \textit{91.62} & \textit{91.46} & \textit{94.38} & \textit{95.32} & \textit{92.17} & \textit{92.78} & \textit{94.27} & \textit{93.67} & \textit{89.75} & \textit{90.62} & \textit{94.27} & \textit{92.76}\\\hline
Summarized Label (SL) & 73.80 & 59.94 & 90.01 & \textbf{94.17} & 70.35 & 90.46 & 90.11 & 87.30 & 60.62 & 75.21 & 85.40 & 79.76\\
EnglishDemos + SL & 82.12 & 64.48 & \underline{92.03} & 92.75 & 81.52 & 90.58 & \textbf{92.25} & 89.30 & 64.28 & 77.71 & 86.23 & 83.02\\
EnglishDemos + Rev. & 83.56 & 70.19 & 92.02 & \underline{93.25} & 81.77 & 90.96 & \underline{91.56} & 84.96 & 65.80 & 81.93 & 86.29 & 83.85\\
TargetDemos & \underline{85.02} & 76.05 & 90.29 & 91.31 & \textbf{87.75} & 90.83 & 89.97 & 89.67 & \underline{79.41} & 82.85 & \underline{89.49} & 86.60\\
TargetDemos + SL & \textbf{85.09} & \textbf{77.65} & 91.70 & 92.75 & 87.16 & 89.59 & 90.79 & \textbf{91.68} & 78.96 & 79.59 & 87.28 & 86.57\\
TargetDemos + Rev. & 83.61 & 70.76 & 91.03 & 93.04 & 87.37 & \underline{91.63} & 91.27 & 90.08 & 78.60 & \textbf{84.48} & \textbf{90.36} & \underline{86.57}\\
TargetDemos + SL + Rev. & 85.00 & \underline{77.07} & \textbf{92.74} & 93.25 & \underline{87.58} & \textbf{91.85} & 90.92 & \underline{90.71} & \textbf{80.23} & \underline{83.30} & 87.44 & \textbf{87.28}\\
\midrule \midrule

\textbf{Topic Classification} & az             & cy             & de             & en             & he             & id             & ro             & sl             & sw             & te             & th             & avg \\ \midrule
Gold Data & \textit{84.85} & \textit{77.03} & \textit{84.44} & \textit{89.99} & \textit{83.65} & \textit{87.77} & \textit{86.53} & \textit{86.04} & \textit{76.64} & \textit{80.29} & \textit{86.94} & \textit{84.01}\\\hline
Summarized Label (SL) & 72.80 & 61.72 & 71.77 & 70.68 & 73.46 & 73.44 & 72.18 & 71.52 & 63.41 & 69.45 & 71.91 & 70.21\\
EnglishDemos + SL & 73.33 & 62.88 & \textbf{78.19} & \textbf{79.65} & \textbf{76.74} & 79.83 & 75.17 & 75.95 & 62.59 & 66.72 & 75.63 & 73.34\\
EnglishDemos + Rev. & \textbf{77.55} & 67.85 & 75.16 & 77.52 & \underline{75.55} & \textbf{80.83} & \underline{76.05} & 80.22 & \textbf{65.71} & 70.30 & \textbf{77.66} & \underline{74.95}\\
TargetDemos & 73.53 & 64.94 & 75.17 & 77.19 & 72.95 & 78.62 & 74.92 & 78.41 & 55.83 & \textbf{76.61} & 72.45 & 72.78\\
TargetDemos + SL & 68.60 & 66.48 & \underline{78.08} & \underline{79.65} & 73.07 & 78.87 & 74.81 & \underline{81.25} & \underline{65.58} & 73.09 & 76.28 & 74.16\\
TargetDemos + Rev. & \underline{74.89} & \textbf{70.46} & 77.58 & 79.19 & 73.21 & \underline{79.88} & \textbf{77.02} & \textbf{81.45} & 60.77 & \underline{74.86} & \underline{76.41} & \textbf{75.06}\\
TargetDemos + SL + Rev. & 72.97 & \underline{69.81} & 77.30 & 77.52 & 73.69 & 76.53 & 72.20 & 80.10 & 62.03 & 72.61 & 75.02 & 73.62\\
 \midrule \midrule

\textbf{Sentiment Analysis} & az             & cy             & de             & en             & he             & id             & ro             & sl             & sw             & te             & th             & avg         \\ \midrule
Gold Data & \textit{71.69} & \textit{58.77} & \textit{65.84} & \textit{80.97} & \textit{82.23} & \textit{90.39} & \textit{90.48} & \textit{83.40} & \textit{75.07} & \textit{82.51} & \textit{80.14} & \textit{78.32}\\\hline
Summarized Label (SL) & 46.59 & 47.16 & 66.83 & 67.99 & 72.45 & 74.54 & 82.41 & 50.69 & 42.02 & \textbf{79.61} & 69.95 & 63.66\\
EnglishDemos + SL & 66.16 & 40.99 & 68.95 & 75.57 & 72.86 & 86.43 & 86.05 & \underline{74.81} & \underline{60.90} & \underline{73.91} & \underline{78.30} & \underline{71.36}\\
EnglishDemos + Rev. & \textbf{71.33} & 35.20 & \textbf{73.08} & 70.48 & \underline{75.28} & \underline{88.29} & 83.58 & 67.66 & \textbf{66.22} & 66.16 & \textbf{79.33} & 70.60\\
TargetDemos & 54.20 & \underline{52.83} & 67.65 & \textbf{78.72} & 66.97 & 83.06 & 82.54 & 55.41 & 57.97 & 61.80 & 74.74 & 66.90\\
TargetDemos + SL & 64.17 & 43.20 & \underline{68.98} & 75.57 & \textbf{75.59} & \textbf{88.90} & 82.64 & 71.75 & 53.51 & 63.98 & 72.76 & 69.19\\
TargetDemos + Rev. & 66.94 & \textbf{53.79} & 65.27 & \underline{77.95} & 69.74 & 84.76 & \textbf{87.11} & 74.40 & 60.55 & 70.14 & 76.56 & \textbf{71.57}\\
TargetDemos + SL + Rev. & \underline{67.23} & 34.06 & 61.59 & 70.48 & 73.83 & 85.99 & \underline{86.48} & \textbf{76.28} & 58.53 & 67.95 & 77.78 & 69.11\\
 \bottomrule
\end{tabular}
}
\caption{Fine-tuning results for the gold and artificial data generated by Llama3-8b on the three tasks. For each configuration we fine-tune ten downstream XLM-R models and report average F1 scores. We bold the best generation strategy in each column and underline the second best. See \ref{sec:lang_abbreviations} for language abbreviations.}\label{tab:llama3-8b-results}
\end{table*}

\begin{table*}[h!]
\resizebox{\textwidth}{!}{
\begin{tabular}{@{}llllllllllll|l@{}}
\toprule
\textbf{Intent Recognition} & az             & cy             & de             & en             & he             & id             & ro             & sl             & sw             & te             & th             & avg         \\ \midrule
Gold Data & \textit{91.62} & \textit{91.46} & \textit{94.38} & \textit{95.32} & \textit{92.17} & \textit{92.78} & \textit{94.27} & \textit{93.67} & \textit{89.75} & \textit{90.62} & \textit{94.27} & \textit{92.76}\\\hline
Summarized Label (SL) & 77.36 & 52.51 & 84.09 & 92.47 & 82.58 & 89.02 & 88.85 & 89.25 & 76.49 & 83.59 & 85.34 & 81.96\\
EnglishDemos + SL & 76.39 & 61.24 & 91.19 & 92.89 & 85.12 & 91.17 & 89.41 & 84.92 & 63.73 & 79.21 & 86.30 & 81.96\\
EnglishDemos + Rev. & 79.84 & 47.33 & \textbf{92.90} & \textbf{92.94} & 80.30 & \underline{91.27} & 86.12 & 82.76 & 66.21 & 77.83 & 87.92 & 80.49\\
TargetDemos & 76.40 & 49.11 & 90.33 & 89.05 & \underline{89.39} & 87.86 & 88.82 & 87.83 & 72.52 & 82.36 & 89.36 & 82.09\\
TargetDemos + SL & \textbf{85.55} & \underline{65.71} & 91.07 & 92.89 & \textbf{89.41} & \textbf{91.91} & \textbf{91.54} & \underline{90.52} & \underline{77.47} & \underline{87.54} & \textbf{90.63} & \underline{86.75}\\
TargetDemos + Rev. & 82.22 & 61.56 & 89.85 & 91.25 & 88.67 & 89.72 & 90.21 & 88.16 & 76.25 & 84.89 & \underline{89.57} & 84.76\\
TargetDemos + SL + Rev. & \underline{84.65} & \textbf{71.88} & \underline{91.79} & \underline{92.94} & 89.26 & 90.96 & \underline{90.79} & \textbf{91.58} & \textbf{79.64} & \textbf{88.01} & 89.08 & \textbf{87.33}\\
\midrule \midrule

\textbf{Topic Classification} & az             & cy             & de             & en             & he             & id             & ro             & sl             & sw             & te             & th             & avg \\ \midrule
Gold Data & \textit{81.94} & \textit{78.61} & \textit{84.91} & \textit{89.99} & \textit{83.98} & \textit{85.33} & \textit{86.38} & \textit{85.83} & \textit{76.71} & \textit{80.31} & \textit{87.51} & \textit{83.77}\\\hline
Summarized Label (SL) & 73.61 & 49.97 & 71.56 & 75.50 & 70.39 & 70.58 & \textbf{75.20} & 72.12 & 53.52 & 65.63 & \underline{77.26} & 68.67\\
EnglishDemos + SL & \textbf{79.39} & 45.36 & \textbf{74.82} & 76.30 & 63.02 & 70.47 & 74.76 & \underline{75.23} & 59.42 & 55.90 & \textbf{78.06} & 68.43\\
EnglishDemos + Rev. & 76.12 & 54.77 & \underline{74.22} & 76.65 & 66.58 & \underline{77.98} & 74.24 & \textbf{75.40} & \textbf{61.98} & 58.16 & 76.47 & 70.23\\
TargetDemos & 76.89 & 44.20 & 70.67 & \textbf{80.15} & 70.57 & 76.92 & 72.74 & 74.64 & 56.27 & 66.00 & 72.76 & 69.26\\
TargetDemos + SL & 69.00 & 38.52 & 70.93 & 76.30 & \textbf{70.95} & 73.20 & 72.49 & 73.26 & 60.34 & 68.53 & 72.73 & 67.84\\
TargetDemos + Rev. & \underline{79.22} & \textbf{64.20} & 73.95 & \underline{79.72} & \underline{70.79} & \textbf{79.22} & 74.14 & 73.73 & 57.38 & \textbf{73.33} & 71.72 & \textbf{72.49}\\
TargetDemos + SL + Rev. & 75.20 & \underline{57.67} & 72.23 & 76.65 & 70.02 & 77.08 & \underline{74.83} & 73.49 & \underline{61.49} & \underline{71.53} & 74.17 & \underline{71.31}\\
 \midrule \midrule

\textbf{Sentiment Analysis} & az             & cy             & de             & en             & he             & id             & ro             & sl             & sw             & te             & th             & avg         \\ \midrule
Gold Data & \textit{71.69} & \textit{58.77} & \textit{65.84} & \textit{80.97} & \textit{82.23} & \textit{90.39} & \textit{90.48} & \textit{83.40} & \textit{75.07} & \textit{82.51} & \textit{80.14} & \textit{78.32}\\\hline
Summarized Label (SL) & \underline{69.86} & 33.88 & 64.50 & 68.52 & \textbf{77.99} & \textbf{88.19} & 83.42 & 68.20 & 56.89 & 70.00 & 69.12 & 68.23\\
EnglishDemos + SL & 68.09 & 33.72 & 60.40 & 69.08 & 62.12 & 77.55 & 81.52 & 70.38 & 34.71 & 61.18 & \underline{76.95} & 63.24\\
EnglishDemos + Rev. & 65.69 & 33.87 & 62.91 & 66.17 & 53.20 & 77.82 & 83.88 & 60.51 & 49.65 & 74.13 & 74.34 & 63.83\\
TargetDemos & 67.98 & 33.70 & 63.70 & \underline{73.19} & 69.23 & 79.08 & \underline{84.67} & \underline{78.02} & \underline{61.31} & 72.83 & 72.44 & 68.74\\
TargetDemos + SL & \textbf{70.77} & 33.63 & 64.27 & 69.08 & \underline{77.84} & 80.34 & 76.22 & 77.98 & 30.57 & 78.56 & 75.27 & 66.78\\
TargetDemos + Rev. & 66.27 & \textbf{49.17} & \textbf{68.97} & \textbf{77.82} & 74.76 & 77.51 & \textbf{85.62} & 76.77 & \textbf{66.30} & \textbf{80.30} & 69.97 & \textbf{72.13}\\
TargetDemos + SL + Rev. & 69.71 & \underline{45.02} & \underline{66.36} & 66.17 & 77.45 & \underline{85.20} & 84.28 & \textbf{79.42} & 31.66 & \underline{78.76} & \textbf{78.08} & \underline{69.28}\\
 \bottomrule
\end{tabular}
}
\caption{Fine-tuning results for the gold and artificial data generated by Gemma3-4b on the three tasks. For each configuration we fine-tune ten downstream XLM-R models and report average F1 scores. We bold the best generation strategy in each column and underline the second best. See \ref{sec:lang_abbreviations} for language abbreviations.}\label{tab:gemma3-4b-results}
\end{table*}

\begin{table*}[h!]
\resizebox{\textwidth}{!}{
\begin{tabular}{@{}llllllllllll|l@{}}
\toprule
\textbf{Intent Recognition} & az             & cy             & de             & en             & he             & id             & ro             & sl             & sw             & te             & th             & avg         \\ \midrule
Gold Data & \textit{91.62} & \textit{91.46} & \textit{94.38} & \textit{95.32} & \textit{92.17} & \textit{92.78} & \textit{94.27} & \textit{93.67} & \textit{89.75} & \textit{90.62} & \textit{94.27} & \textit{92.76}\\\hline
Summarized Label (SL) & 76.28 & 74.84 & 87.96 & 92.42 & 85.06 & 89.82 & 88.81 & 88.05 & 80.05 & 84.88 & 87.98 & 85.10\\
EnglishDemos + SL & 86.66 & 74.02 & 92.51 & \textbf{93.02} & 85.58 & 92.91 & 87.14 & 81.71 & 81.58 & 77.39 & 86.45 & 85.36\\
EnglishDemos + Rev. & 87.15 & \underline{77.91} & \underline{92.64} & 92.93 & \textbf{91.32} & \underline{93.28} & \textbf{92.79} & \underline{91.69} & \textbf{82.38} & 87.30 & 85.80 & \underline{88.65}\\
TargetDemos & 73.97 & 66.66 & 88.79 & 89.56 & 89.78 & 90.52 & 91.24 & \textbf{91.90} & 80.61 & 83.07 & 85.68 & 84.71\\
TargetDemos + SL & \underline{87.17} & 70.97 & 91.52 & \underline{93.02} & 88.97 & 91.79 & \underline{91.49} & 90.55 & 79.51 & \textbf{88.38} & \underline{88.95} & 87.48\\
TargetDemos + Rev. & 86.08 & 76.14 & 91.60 & 92.14 & \underline{90.29} & 91.15 & 91.46 & 91.64 & 81.28 & 87.18 & 86.99 & 87.81\\
TargetDemos + SL + Rev. & \textbf{88.81} & \textbf{79.35} & \textbf{93.16} & 92.93 & 88.80 & \textbf{93.65} & 90.15 & 88.48 & \underline{82.11} & \underline{87.84} & \textbf{90.19} & \textbf{88.68}\\
\midrule \midrule

\textbf{Topic Classification} & az             & cy             & de             & en             & he             & id             & ro             & sl             & sw             & te             & th             & avg \\ \midrule
Gold Data & \textit{81.94} & \textit{78.61} & \textit{84.91} & \textit{89.99} & \textit{83.98} & \textit{85.33} & \textit{86.38} & \textit{85.83} & \textit{76.71} & \textit{80.31} & \textit{87.51} & \textit{83.77}\\\hline
Summarized Label (SL) & 73.92 & 59.52 & 72.63 & 72.08 & 65.63 & 69.27 & \underline{73.39} & 71.80 & 62.16 & 64.99 & 72.36 & 68.89\\
EnglishDemos + SL & \underline{77.12} & \underline{64.35} & 73.43 & 75.12 & \textbf{71.09} & 74.44 & 69.20 & 70.82 & 62.26 & \textbf{70.28} & 75.66 & 71.25\\
EnglishDemos + Rev. & 76.50 & \textbf{65.93} & 73.81 & 74.49 & \underline{69.92} & 72.05 & 72.03 & \textbf{74.83} & \textbf{67.28} & \underline{70.22} & \underline{76.62} & \textbf{72.15}\\
TargetDemos & 75.51 & 49.63 & 71.54 & \underline{76.93} & 68.93 & 74.21 & 69.17 & 73.25 & 63.20 & 69.78 & 73.79 & 69.63\\
TargetDemos + SL & 73.64 & 57.87 & 74.30 & 75.12 & 65.88 & 74.55 & 70.11 & 74.57 & 58.53 & 66.90 & 76.15 & 69.78\\
TargetDemos + Rev. & \textbf{77.36} & 62.55 & \underline{74.69} & \textbf{77.72} & 65.20 & \textbf{75.49} & \textbf{75.80} & \underline{74.70} & \underline{67.22} & 67.21 & 72.57 & \underline{71.86}\\
TargetDemos + SL + Rev. & 74.52 & 62.14 & \textbf{75.47} & 74.49 & 68.34 & \underline{75.19} & 70.99 & 71.86 & 59.22 & 69.33 & \textbf{76.78} & 70.76\\
 \midrule \midrule

\textbf{Sentiment Analysis} & az             & cy             & de             & en             & he             & id             & ro             & sl             & sw             & te             & th             & avg         \\ \midrule
Gold Data & \textit{71.69} & \textit{58.77} & \textit{65.84} & \textit{80.97} & \textit{82.23} & \textit{90.39} & \textit{90.48} & \textit{83.40} & \textit{75.07} & \textit{82.51} & \textit{80.14} & \textit{78.32}\\\hline
Summarized Label (SL) & 63.46 & 35.38 & 66.55 & 65.80 & \textbf{78.04} & 71.52 & \textbf{81.42} & 41.84 & 53.19 & 69.79 & \textbf{75.28} & 63.84\\
EnglishDemos + SL & \textbf{68.44} & 35.05 & 62.93 & 62.39 & 72.65 & 83.69 & 78.81 & 46.55 & 33.32 & 70.32 & 74.15 & 62.57\\
EnglishDemos + Rev. & \underline{65.44} & 36.91 & \underline{67.58} & 60.59 & 68.62 & 71.72 & 78.98 & 58.95 & 40.87 & 65.87 & 71.00 & 62.41\\
TargetDemos & 63.33 & \underline{46.73} & \textbf{68.15} & \underline{75.95} & 73.87 & 87.41 & 72.86 & 58.32 & \textbf{61.71} & 53.56 & 69.66 & 66.51\\
TargetDemos + SL & 61.89 & 33.87 & 65.02 & 62.39 & 78.01 & \underline{89.73} & \underline{80.72} & 65.70 & 49.77 & \underline{70.90} & \underline{74.50} & \underline{66.59}\\
TargetDemos + Rev. & 63.34 & \textbf{60.60} & 65.99 & \textbf{77.92} & \underline{78.02} & \textbf{90.13} & 74.99 & \underline{73.78} & \underline{53.35} & 62.75 & 71.14 & \textbf{70.18}\\
TargetDemos + SL + Rev. & 62.41 & 34.92 & 63.49 & 60.59 & 73.55 & 89.39 & 69.89 & \textbf{75.20} & 53.22 & \textbf{75.60} & 72.49 & 66.43\\
 \bottomrule
\end{tabular}
}
\caption{Fine-tuning results for the gold and artificial data generated by Gemma3-27b on the three tasks. For each configuration we fine-tune ten downstream XLM-R models and report average F1 scores. We bold the best generation strategy in each column and underline the second best. See \ref{sec:lang_abbreviations} for language abbreviations.}\label{tab:gemma3-27b-results}
\end{table*}

\begin{table*}[h!]
\resizebox{\textwidth}{!}{
\begin{tabular}{@{}llllllllllll|l@{}}
\toprule
\textbf{Intent Recognition} & az             & cy             & de             & en             & he             & id             & ro             & sl             & sw             & te             & th             & avg         \\ \midrule
Gold Data & \textit{91.62} & \textit{91.46} & \textit{94.38} & \textit{95.32} & \textit{92.17} & \textit{92.78} & \textit{94.27} & \textit{93.67} & \textit{89.75} & \textit{90.62} & \textit{94.27} & \textit{92.76}\\\hline
Summarized Label (SL) & \underline{85.10} & 65.58 & 90.19 & 90.61 & 76.16 & 90.69 & 88.28 & 89.27 & 70.23 & 83.71 & 86.66 & 83.31\\
EnglishDemos + SL & 83.82 & 76.08 & \underline{92.60} & 92.50 & 81.88 & 90.56 & 88.08 & 88.02 & 74.64 & 84.32 & 86.05 & 85.32\\
EnglishDemos + Rev. & 84.33 & 76.31 & 92.02 & \textbf{92.99} & 83.02 & 91.41 & 91.55 & 89.31 & 73.10 & \underline{85.42} & 87.30 & 86.07\\
TargetDemos & 84.24 & \textbf{81.98} & 91.97 & 92.26 & \textbf{88.76} & \underline{91.65} & 90.40 & 88.13 & 80.03 & 85.16 & \underline{88.79} & 87.58\\
TargetDemos + SL & 85.00 & 79.89 & \textbf{92.79} & 92.50 & 88.65 & 90.79 & \textbf{92.52} & \textbf{90.06} & \textbf{80.46} & 84.94 & 88.20 & \textbf{87.80}\\
TargetDemos + Rev. & \textbf{86.09} & 79.66 & 91.53 & 92.96 & 88.46 & \textbf{91.74} & \underline{91.78} & \underline{89.77} & \underline{80.38} & \textbf{86.06} & 86.92 & \underline{87.76}\\
TargetDemos + SL + Rev. & 85.06 & \underline{80.28} & 92.56 & \underline{92.99} & \underline{88.74} & 90.92 & 90.66 & 89.03 & 79.77 & 85.11 & \textbf{89.40} & 87.68\\
\midrule \midrule

\textbf{Topic Classification} & az             & cy             & de             & en             & he             & id             & ro             & sl             & sw             & te             & th             & avg \\ \midrule
Gold Data & \textit{84.85} & \textit{77.03} & \textit{84.44} & \textit{89.99} & \textit{83.65} & \textit{87.77} & \textit{86.53} & \textit{86.04} & \textit{76.64} & \textit{80.29} & \textit{86.94} & \textit{84.01}\\\hline
Summarized Label (SL) & 77.69 & 56.36 & 75.68 & 73.28 & 65.97 & 71.44 & 72.02 & 74.87 & 60.15 & 70.45 & 75.28 & 70.29\\
EnglishDemos + SL & 70.73 & 70.75 & 73.25 & \textbf{81.92} & \underline{76.05} & \underline{79.52} & 73.56 & 78.85 & 68.80 & 72.03 & 77.71 & 74.83\\
EnglishDemos + Rev. & 75.84 & 69.39 & \textbf{78.27} & 79.30 & 75.88 & 78.56 & \underline{76.94} & 79.23 & \textbf{70.71} & 73.40 & \textbf{80.70} & 76.20\\
TargetDemos & \textbf{79.55} & \textbf{71.75} & 75.51 & 80.03 & 75.90 & 78.88 & \textbf{78.49} & \textbf{81.55} & 68.93 & 73.33 & 76.96 & \textbf{76.44}\\
TargetDemos + SL & 75.63 & 70.47 & 76.33 & \underline{81.92} & 75.37 & 78.63 & 76.70 & 79.13 & 68.78 & \textbf{74.75} & 79.13 & 76.08\\
TargetDemos + Rev. & \underline{77.74} & \underline{70.86} & \underline{77.25} & 80.78 & 76.04 & \textbf{79.63} & 75.79 & \underline{80.89} & 67.98 & \underline{73.61} & 79.66 & \underline{76.39}\\
TargetDemos + SL + Rev. & 76.26 & 70.05 & 77.20 & 79.30 & \textbf{76.89} & 77.58 & 76.44 & 79.46 & \underline{69.47} & 73.10 & \underline{80.23} & 76.00\\
 \midrule \midrule

\textbf{Sentiment Analysis} & az             & cy             & de             & en             & he             & id             & ro             & sl             & sw             & te             & th             & avg         \\ \midrule
Gold Data & \textit{71.69} & \textit{58.77} & \textit{65.84} & \textit{80.97} & \textit{82.23} & \textit{90.39} & \textit{90.48} & \textit{83.40} & \textit{75.07} & \textit{82.51} & \textit{80.14} & \textit{78.32}\\\hline
Summarized Label (SL) & 66.72 & 44.79 & 65.65 & 69.42 & 74.56 & 67.70 & \underline{84.06} & 40.22 & 33.70 & 46.94 & 75.42 & 60.83\\
EnglishDemos + SL & 68.15 & \underline{60.09} & \textbf{71.34} & 73.90 & 72.78 & 85.70 & 82.94 & 46.36 & 45.58 & 51.12 & \underline{79.82} & 67.07\\
EnglishDemos + Rev. & 62.86 & 60.09 & \underline{67.83} & 65.04 & 72.99 & 85.96 & 77.42 & 52.71 & 46.56 & 58.04 & 79.49 & 66.27\\
TargetDemos & 71.32 & \textbf{60.16} & 60.84 & \textbf{78.62} & 71.44 & 86.41 & 81.83 & \underline{73.14} & \textbf{67.08} & 75.51 & 75.72 & \textbf{72.92}\\
TargetDemos + SL & \textbf{73.43} & 59.41 & 63.21 & 73.90 & 70.45 & \textbf{89.00} & 79.47 & 71.11 & 59.17 & \textbf{79.00} & \textbf{79.99} & 72.56\\
TargetDemos + Rev. & \underline{71.52} & 54.08 & 67.13 & \underline{75.03} & \underline{76.32} & \underline{88.30} & \textbf{85.62} & 64.93 & \underline{61.18} & 76.22 & 78.72 & \underline{72.64}\\
TargetDemos + SL + Rev. & 69.40 & 34.58 & 63.15 & 65.04 & \textbf{77.02} & 86.69 & 78.29 & \textbf{74.88} & 57.94 & \underline{77.64} & 79.21 & 69.44\\
 \bottomrule
\end{tabular}
}
\caption{Fine-tuning results for the gold and artificial data generated by Llama3-70b on the three tasks. For each configuration we fine-tune ten downstream XLM-R models and report average F1 scores. We bold the best generation strategy in each column and underline the second best. See \ref{sec:lang_abbreviations} for language abbreviations.}\label{tab:llama3-70b-results}
\end{table*}

\subsection{Data Diversity}\label{app:data-diversity}

Table \ref{tab:ngram-diversity} shows the results of computing the n-gram diversity for the original gold and the data generated with the target language demonstrations and revision. The results are aggregated across all languages.

\begin{table*}[]
\centering
\small
\begin{tabular}{l|l|l|l|l|l}
\toprule
 & \multicolumn{1}{c|}{Gold} & \multicolumn{1}{c|}{Gemma3-4b} & \multicolumn{1}{c|}{Llama3-8b} & \multicolumn{1}{c|}{Gemma3-27b} & \multicolumn{1}{c}{Llama3-70b} \\ \midrule
Intent Recognition   & \textit{2.475}                    & 2.913                         & 2.414                         & \textbf{3.033}                 & 2.412                          \\
Topic Classification & \textbf{\textit{3.377}}           & 3.242                         & 2.970                         & 3.292                          & 2.890                          \\
Sentiment Analysis   & \textit{2.614}                    & 3.376                         & 3.045                         & \textbf{3.437}                 & 2.921                      \\
\bottomrule
\end{tabular}\caption{N-gram diversity of the gold and generated data aggregated across all languages. The generation strategy includes demonstrations in the target language and revision. Numbers in bold indicate the largest diversity per task.}\label{tab:ngram-diversity}
\end{table*}

\end{document}